\pdfoutput=1
\documentclass[11pt,twoside]{article}

\newcommand*\TitleRunning{Online Saddle Point Problem and Online Convex-Concave Optimization}
\newcommand*\AuthorRunning{Qing-xin Meng and Jian-wei Liu}
\newcommand*\Keywords{Online Saddle Point Problem, Online Convex-Concave Games, Implicit Updates, Duality Gap, Optimistic Online Learning}
\usepackage{arxiv}
\title{Online Saddle Point Problem and Online Convex-Concave Optimization}

\author{\sc Qing-xin Meng
\orcid{0000-0003-4014-7405}
\\\small\email{qingxin6174@gmail.com}
\and\sc Jian-wei Liu
\\\small\email{liujw@cup.edu.cn}
}

\addr{Department of Automation, College of Information Science and Engineering \\
China University of Petroleum, Beijing, Beijing 102249, China}

\begin{document}
\maketitle\thispagestyle{firstpage}

\begin{abstract}
Centered around solving the Online Saddle Point problem, this paper introduces the Online Convex-Concave Optimization~(OCCO) framework, which involves a sequence of two-player time-varying convex-concave games. 
We propose the generalized duality gap~(Dual-Gap) as the performance metric and establish the parallel relationship between OCCO with Dual-Gap and Online Convex Optimization~(OCO) with regret. 
To demonstrate the natural extension of OCCO from OCO, we develop two algorithms, the implicit online mirror descent-ascent and its optimistic variant. 
Analysis reveals that their duality gaps share similar expression forms with the corresponding dynamic regrets arising from implicit updates in OCO. 
Empirical results further substantiate the effectiveness of our algorithms.
Simultaneously, we unveil that the dynamic Nash equilibrium regret, which was initially introduced in a recent paper, has inherent defects. 
\end{abstract}


\section{Introduction}

Consider the Online Saddle Point~\citep[OSP,][]{cardoso2018online} problem which involves a sequence of two-player time-varying convex-concave games. 
Specifically, in each round, players~1 and~2 jointly select a pair of strategies wherein player~1 minimizes his payoff, while player~2 maximizes his payoff. 
Both players make decisions without foreknowledge of the current and future payoff functions. 
Once the strategy pair is determined, the environment reveals a convex-concave payoff function. 
No additional assumptions are imposed on the environment, thereby allowing for potential regularity or even adversarial behavior. 
The goal is to provide players with decision-making algorithms that closely approximate Nash equilibriums.

Precisely, there is a sequence of payoff functions $f_t\left(x,y\right)$, for $t=1, 2, \cdots$, where $f_t$ is continuous and $f_t\left(\,\cdot\,,y\right)$ is convex in $X$ for every $y\in Y$ and $f_t\left(x, \,\cdot\,\right)$ is concave in $Y$ for every $x\in X$, $X$ and $Y$ are compact convex sets in normed linear spaces $\left(\mathscr{X}, \left\lVert\cdot\right\rVert_\mathscr{X}\right)$ and $\left(\mathscr{Y}, \left\lVert\cdot\right\rVert_\mathscr{Y}\right)$, respectively. 
We may omit norm subscripts without causing ambiguity, for example by explicitly determining the norm according to the space to which the element belongs. 
\citet[][Theorem~3]{kakutani1941generalization} 
guarantees the existence of a Nash equilibrium, that is, there exists a saddle point $\left(x_t^*, y_t^*\right)\in X\times Y$ such that 
\begin{equation*}
\begin{aligned}
\min_{x\in X}\max_{y\in Y}f_t\left(x,y\right)=f_t\left(x_t^*, y_t^*\right)=\max_{y\in Y}\min_{x\in X}f_t\left(x,y\right). 
\end{aligned}
\end{equation*}
Equivalent results in minimax theory can be found in \citet{frenk2004minimax}. 

Notice that the OSP problem is an extended application of Online Convex Optimization~\citep[OCO,][]{zinkevich2003online} involving two players. 
Therefore, identifying online scenarios suitable for OSP becomes straightforward. Examples include dynamic pricing and online advertising auctions, which fall under the broader category of budget-reward trade-offs. 

Most conventional literature on offline convex-concave optimization employs the duality gap as the performance metric, which is given by $f(x_t,y^*)-f(x^*,y_t)$. 
In the online setting, we consider the following \emph{generalized duality gap}: 
\begin{equation}
\label{def:Dual-Gap}
\begin{aligned}
\textrm{Dual-Gap}\ \{(u_t,v_t)\}_{t=1:T}\coloneq\sum_{t=1}^T \big(f_t\left(x_t, v_t\right)-f_t\left(u_t, y_t\right)\big),
\end{aligned}
\end{equation}
where $\{(u_t,v_t)\in X\times Y\}_{t=1:T}$ represents an arbitrary competitor sequence. 
For the OSP problem, it suffices to choose $(u_t,v_t)=(x_t^*, y_t^*)$. 
This definition draws parallels with the concept of dynamic regret in non-stationary OCO. 
In adversarial scenarios, considering non-saddle point sequences as competitors can provide advantages by avoiding vacuous bounds. 
The generalized duality gap offers an additional benefit by accommodating other performance metrics, such as \citep{zhang2022noregret}: 
\begin{equation}
\label{eq:dual-gap}
\begin{aligned}
\sum_{t=1}^T\Big(\max_{y\in Y} f_t\left(x_t, y\right)-\min_{x\in X}f_t\left(x, y_t\right)\Big).
\end{aligned}
\end{equation}
which constitutes the time-varying counterpart of the saddle-point residual~\citep{gao2019increasing,grand2021conic}. 

We now formally present Online Convex-Concave Optimization~(OCCO) with full information\footnote{Here, ``\emph{full information}'' refers to each game round revealing players complete information of the underlying game. This is stronger than the full information in the OCO framework, where player~1 observes $f_t(\,\cdot\,,y_t)$ and player~2 observes $f_t(x_t,\,\cdot\,)$. }: 
In round $t$, 
\begin{itemize}[itemsep=1pt,topsep=2pt]
\item players~1 and~2 jointly select a pair of strategies $\left(x_t,y_t\right)\in X\times Y$, 
\item the environment feeds back a continuous convex-concave function $f_t$. 
\end{itemize}
The aim of this game is to minimize the generalized duality gap with respect to arbitrary competitor sequence $\{(u_t,v_t)\}_{t=1:T}$. 
OCCO encompasses the OSP problem and naturally extends OCO. 
In fact, OCCO can be reduced to OCO if we constrain the action set of one player to a singleton. 
In such cases, the generalized duality gap reduces to dynamic regret, specifically, $f_t\left(x_t, y\right)-f_t\left(u_t, y\right)$ if $Y=\{y\}$, or $-f_t\left(x, y_t\right)-(-f_t)\left(x, v_t\right)$ if $X=\{x\}$. 
This leads to the question: Can we address OCCO similarly to how we solve OCO? 
Drawing inspiration from the success of implicit updates in OCO, we aim to demonstrate the following idea:
\begin{itemize}
\item[]\emph{The efficacy of implicit updates for solving OCO with dynamic regret can be extended to solve OCCO with the generalized duality gap. }
\end{itemize}

This paper substantiates the aforementioned notion. 
To achieve this, we propose two algorithms: the Implicit Online Mirror Descent-Ascent~(IOMDA) and its optimistic variant, the Optimistic Implicit Online Mirror Descent-Ascent~(OptIOMDA). 
Our analysis yields the following results: 
\begin{itemize}[itemsep=1pt,topsep=0pt]
\item IOMDA with adaptive learning rates achieves that 
\begin{equation*}
\begin{aligned}
\textrm{Dual-Gap}\ \{(u_t,v_t)\}_{t=1:T}\leqslant O\left(\min\bigg\{\sum_{t=1}^{T}\rho\left(f_{t}, f_{t-1}\right), P+\sqrt{(1+P)T}\bigg\}\right),
\end{aligned}
\end{equation*}  
\item and OptIOMDA with adaptive learning rates guarantees that 
\begin{equation*}
\begin{aligned}
\textrm{Dual-Gap}\ \{(u_t,v_t)\}_{t=1:T}\leqslant O\left(\min\bigg\{\sum_{t=1}^{T}\rho\left(f_{t}, h_{t}\right), \sqrt{(1+P)T}\bigg\}\right),
\end{aligned}
\end{equation*}
\end{itemize}
where $\rho\left(f, g\right)=\max_{x\in X,\,y\in Y}\left\lvert f(x,y)-g(x,y)\right\rvert$ measures the distance between $f$ and $g$, $h_{t}$ represents an arbitrary predictor, and $P$ stands as an upper bound of $P_T= \sum_{t=1}^T \big(\lVert u_t-u_{t-1}\rVert+\lVert v_t-v_{t-1}\rVert\big)$, which is the path length of the competitor sequence. 
By drawing from the work of \citet{Campolongo2021closer}, we demonstrate a profound parallelism between OCCO with Dual-Gap and OCO with regret.
To the best of our knowledge, we are the first to offer generalized duality gap guarantees for both benign and hostile OCCO scenarios. 

Furthermore, in a recent work by \citet{zhang2022noregret}, they introduce an alternative performance metric called dynamic Nash equilibrium regret~(NE-regret), defined as:
\begin{equation}
\label{eq:NE}
\begin{aligned}
\textrm{NE-regret}_{T}
\coloneq\left\lvert \sum_{t=1}^T f_t\left(x_t, y_t\right)-\sum_{t=1}^T f_t\left(x_t^*, y_t^*\right)\right\rvert.
\end{aligned}
\end{equation}
However, it is worth noting that NE-regret is not a good metric for learning in time-varying games. 
In fact, having sublinear NE-regret does not guarantee that players' strategies asymptotically converge to Nash equilibriums. 
For instance, consider a scenario where in half of the iterates, the value is much larger than the value at NE, while in the other half of the iterates, the opposite holds. 
This result in a small NE-regret, but it wouldn't indicate how closely the value of each iterate approaches NE. 
In conclusion, the environment can intentionally deceive players, causing elimination inside the absolute value of NE-regret. 
We did not specifically devise experiments to capitalize on this vulnerability, but the oscillations evident in \cref{fig:case31} indeed demonstrate an elimination process. 
Discussion on NE-regret can be found in \cref{app:NE-regret}. 

\section{Related Work}

In this section, we review two lines of work most related to ours: online saddle point optimization and implicit updates.
We outline existing results and show similarities and differences. 

\subsection{Online Saddle Point Optimization}

Numerous works focus on offline saddle point optimization, while the literature on Online Saddle Point~(OSP) problem remains scarce. 
\citet{nguyen2017exploiting,nguyen2019exploiting} study online matrix games, where the payoff function is a time-varying matrix. 
However, their work does not investigate convergence to a Nash equilibrium, only showing that the sum of the individual regrets of the two players is sublinear. 
\citet{cardoso2018online} first consider the OSP problem explicitly and introduce the saddle-point regret~(SP-regret): 
\begin{equation}
\label{eq:sp-regret}
\begin{aligned}
\left\lvert \sum_{t=1}^T f_t\left(x_t, y_t\right)-\min_{x\in X}\max_{y\in Y}\sum_{t=1}^T f_t\left(x,y\right)\right\rvert. 
\end{aligned}
\end{equation}
In a subsequent publication, \citet{cardoso2019competing} rename 
\cref{eq:sp-regret} 
as Nash equilibrium regret and propose an FTRL-like algorithm that achieves an $\widetilde{O}(\sqrt{T})$ Nash equilibrium regret. 
They also prove that no algorithm can simultaneously achieve sublinear Nash equilibrium regret and sublinear individual regret for both players. 
Building upon this, \citet{zhang2022noregret} formalize the concept of dynamic Nash equilibrium regret by rationalizing the SP-regret, moving the minimax operation inside the summation, and defining it as \cref{eq:NE}. 
Additionally, they also investigate the duality gap defined by \cref{eq:dual-gap}. 
They solve the OSP problem by converting dynamic Nash Equilibrium regret and duality gap into dynamic regret, and stacking OCO algorithms. 

This paper proposes an innovative method to solve the OSP problem by introducing the OCCO framework and the generalized duality gap. 
In contrast to the methodology proposed by \citet{zhang2022noregret}, which involves downgrading the OSP problem to OCO, our method highlights a profound parallelism between OCCO and OCO. 
Meanwhile, we also assert that the NE-regret introduced by \citet{zhang2022noregret} is a flawed measure. 

\subsection{Implicit Updates}

Implicit updates are known as proximal methods in the optimization literature. 
This technique can be traced back to the work by \citet{Moreau1965Proximite} and was introduced to the online learning field by \citet{kivinen1997exponentiated}. 
In OCO, implicit updates tackle convex optimizations directly instead of dealing with surrogate linear optimizations. 
\citet{Campolongo2020Temporal} analyze the theoretical advantages of online implicit updates, showcasing that the static regret upper bound not only provides a worst-case guarantee but can also be tightened based on the temporal variability of loss functions. 
In a subsequent publication, \citet{Campolongo2021closer} expand their previous results to dynamic regret: Implicit Online Mirror Descent~(IOMD) enjoys an $O\big(\min\{V_T, \sqrt{(1+P)T}\}\big)$ dynamic regret, where $V_{T}=\sum_{t=1}^{T}\rho\left(\ell_{t}, \ell_{t-1}\right)$ denotes the temporal variability of the loss functions~\citep{besbes2015nonstationary}, and $\ell_t$ is the convex loss. 
Furthermore, \citet{scroccaro2022adaptive} analyze an optimistic version of IOMD, extending the results of \citet{Campolongo2021closer}. 
They demonstrate that the optimistic IOMD achieves the dynamic regret of $O\big(\min\{V'_T, \sqrt{(1+P)T}\}\big)$, where $V'_{T}=\sum_{t=1}^{T}\rho\left(\ell_{t}, h_{t}\right)$ denotes the cumulative distance from loss functions to an arbitrary predictor sequence, and $h_{t}$ is the predictor. 

This paper leverages the groundwork laid by \citet{Campolongo2021closer} to expand the application of implicit updates from OCO to OCCO. 

\section{Preliminaries}

Throughout this paper, we use the abbreviated notation $1:n$ to represent the set $\{1,2,\cdots,n\}$, and define $(\,\cdot\,)_+\coloneq\max(\,\cdot\,,0)$. 
For asymptotic upper bounds, we employ big-$O$ notation, while $\widetilde{O}$ is utilized to conceal poly-logarithmic factors. 

Let's take player~1 as an illustration. 
$\mathscr{X}$ is a normed linear space over $\mathbb{R}$ with norm $\left\lVert\,\cdot\,\right\rVert_{\mathscr{X}}$, and $\mathscr{X}^*$ is the dual space. 
Denote by $\langle\,\cdot\,,\cdot\,\rangle\colon \mathscr{X}^*\times \mathscr{X}\rightarrow\mathbb{R}$ the canonical dual pairing. 

Define the Bregman divergence w.r.t. a proper function $\varphi$ as 
\begin{equation*}
B_{\varphi}\left(x, z\right)\coloneqq\varphi\left(x\right)+\varphi^\star\left(z\right)-\left\langle z, x\right\rangle, \quad\forall\left(x, z\right)\in \mathscr{X}\times \mathscr{X}^*, 
\end{equation*}
where $\varphi^\star$ is the convex conjugate of $\varphi$ given by $\varphi^\star\left(z\right)\coloneqq\sup_{x\in \mathscr{X}} \langle z, x\rangle-\varphi\left(x\right)$. 
This definition with symmetrical aesthetic is based on subdifferentiation, that is, if $\varphi$ is convex, then $\forall z\in\partial\varphi\left(y\right)$, $\varphi\left(x\right)-\varphi\left(y\right)-\left\langle z, x-y\right\rangle=\varphi\left(x\right)+\varphi^\star\left(z\right)-\left\langle z, x\right\rangle$. 
By Fenchel-Young inequality, $B_{\varphi}\left(x, z\right)\geqslant 0$, and equality holds iff $z$ is the subgradient of $\varphi$ at $x$. 
We use the superscripted notation $x^\varphi$ to abbreviate one subgradient of $\varphi$ at $x$. 
Directly applying the definition of Bregman divergence yields $B_{\varphi}\left(x, y^\varphi\right)+B_{\varphi}\left(y, z\right)-B_{\varphi}\left(x, z\right)=\left\langle z-y^\varphi, x-y\right\rangle$. 
A similar version can be found in \citet[][Lemma~3.1]{chen1993convergence}. 

$\varphi$ is $\mu$-strongly convex if $B_{\varphi}\left(x, y^\varphi\right)\geqslant\frac{\mu}{2}\left\lVert x-y\right\rVert^2$, $\forall x\in \mathscr{X}$, $\forall y^\varphi\in\partial\varphi\left(y\right)$. 

$B_\varphi$ is $L$-Lipschitz on $X$ w.r.t. the first variable, if $\exists L$, such that 
\begin{equation*}
\left\lvert B_{\varphi}\left(x_1, y^\varphi\right)-B_{\varphi}\left(x_2, y^\varphi\right)\right\rvert\leqslant L\left\lVert x_1-x_2\right\rVert,
\quad\forall y\in\mathscr{X},
\quad\forall y^\varphi\in\partial\varphi(y),
\quad\forall x_1, x_2\in\mathscr{X}.
\end{equation*}

\section{Main Results}

Begin by appending an additional assumption to OCCO as follows. 

\begin{assumption}
\label{ass:gradient-bounded}
Both $f_t(\,\cdot\,,y)$ and $-f_t(x,\,\cdot\,)$ are gradient-bounded for all $x\in X$ and $y\in Y$, that is, $\exists G_X, G_Y<+\infty$, such that $\left\lVert\partial_x f_t(x,y)\right\rVert\leqslant G_X$ and $\left\lVert\partial_y (-f_t)(x,y)\right\rVert\leqslant G_Y$, $\forall x\in X$, $\forall y\in Y$. 
\end{assumption}

Note that one of the sufficient conditions for the existence of a saddle point is the compactness of the action sets $X$ and $Y$, which consequently implies their boundedness. 
However, rather than directly assuming boundedness, we opt to impose a constraint on the supremum of $B_\phi$ and $B_\psi$. 
This aspect will be elaborated on in \cref{lem:IOMDA}. 
%

In \cref{sec:IOMDA}, we introduce the Implicit Online Mirror Descent-Ascent and establish its Dual-Gap upper bound. 
Moving on to \cref{sec:OptIOMDA}, we elaborate into Optimistic Implicit Online Mirror Descent-Ascent and provide its Dual-Gap upper bound. 
Detailed proofs are deferred to \cref{sec:analysis}.


\subsection{Implicit Online Mirror Descent-Ascent}
\label{sec:IOMDA}

Inspired by the following Implicit Online Mirror Descent~(IOMD): 
\begin{equation}
\label{eq:IOMD}
\begin{aligned}
x_{t+1}=\arg\min_{x\in X}\ell_t(x)+\frac{1}{\eta_t}B_{\phi}\big(x, x_t^\phi\big),
\end{aligned}
\end{equation}
where $\ell_t$ is a convex loss, we consider the following update: 
\begin{equation}
\label{eq:IOMDA}
\begin{aligned}
F_t(x,y)&= f_t(x,y)+\frac{1}{\eta_t}B_{\phi}\big(x, x_t^\phi\big)-\frac{1}{\eta_t}B_{\psi}\big(y, y_t^\psi\big), \\
x_{t+1}&=\arg\min_{x\in X}\max_{y\in Y}F_t(x,y), \\
y_{t+1}&=\arg\max_{y\in Y}\min_{x\in X}F_t(x,y), 
\end{aligned}
\end{equation}
where $\phi\colon\mathscr{X}\rightarrow\mathbb{R}$ and $\psi\colon\mathscr{Y}\rightarrow\mathbb{R}$ are all $1$-strongly convex functions, $x_{t}^\phi\in\partial\phi\left(x_{t}\right)$, $y_{t}^\psi\in\partial\psi\left(y_{t}\right)$, and $\eta_t>0$ represents the learning rate. 
We call \cref{eq:IOMDA} the Implicit Online Mirror Descent-Ascent~(IOMDA). 
In words, IOMDA updates the saddle point of $F_t$ in $X\times Y$ as the prediction for the next round. 

\begin{theorem}[Dual-Gap bound for IOMDA]
\label{lem:IOMDA}
Let $B_\phi$ and $B_\psi$ be $L_\phi$-Lipschitz on $X$ and $L_\psi$-Lipschitz on $Y$ respectively w.r.t. their first variables, and let $D_\phi^2$ and $D_\psi^2$ be the supremum of $B_\phi$ and $B_\psi$, respectively. 
IOMDA with non-increasing learning rates guarantees that 
\begin{equation*}
\begin{aligned}
\textrm{Dual-Gap}\ \{(u_t,v_t)\}_{t=1:T}
\leqslant \frac{C+L P_T}{\eta_T} + \sum_{t=1}^{T}\delta_t, 
\end{aligned}
\end{equation*}
where $C=2D\left(D+\sqrt{2}L\right)$, $D=\max\{D_\phi, D_\psi\}$, $L=\max\{L_\phi, L_\psi\}$, $P_T= \sum_{t=1}^T \big(\lVert u_t-u_{t-1}\rVert+\lVert v_t-v_{t-1}\rVert\big)$, and 
\begin{equation}
\label{eq:delta}
\begin{aligned}
\delta_t=\ &f_t(x_t,v_t)-f_t(x_{t+1}, v_{t+1})+f_t(u_{t+1}, y_{t+1})-f_t(u_t, y_t) \\
&-\big[B_{\phi}\big(x_{t+1}, x_{t}^\phi\big)+B_{\psi}\big(y_{t+1}, y_{t}^\psi\big)\big]/\eta_t.
\end{aligned}
\end{equation}
\end{theorem}


Next, we use an adaptive strategy to set learning rates. 
We explored the adaptive trick adopted in \citet[][Theorem~5.1]{Campolongo2021closer}, where for an upper bound with the expression of 
\[\frac{C+L P_T}{\eta_T}+\sum_{t=1}^T\delta_t,\quad C\in\mathbb{R}_+,\quad\delta_t\geqslant 0,\quad\forall t,\]
the learning rate is set to satisfy $(C + L P)/\eta_t=\sum_{\tau=1}^{t-1}\delta_{\tau}$ with $P$ being the upper bound of $P_T$. 
However, this adaptive trick cannot be applied directly because $\delta_t$ in \cref{lem:IOMDA} may be negative. 
Another minor annoyance is that $\delta_t$ contains information about the future, that is, $\delta_t$ depends on $u_{t+1}$ and $v_{t+1}$, and the consequence is that the current learning rate $\eta_t$ depends on unknown $u_{t}$ and $v_{t}$. 
The following theorem states that the adaptive trick can be satisfied by relaxing the upper bound.

\begin{theorem}[Dual-Gap bound for IOMDA with Adaptive Learning Rates]
\label{thm:IOMDA}
Under the assumptions of \cref{lem:IOMDA} and let $f_t$ satisfy \cref{ass:gradient-bounded}, let $P$ be a preset upper bound of $P_T$. 
IOMDA with $\left(C+L P\right)/\eta_t=\epsilon + \sum_{\tau=1}^{t-2}\Delta_{\tau}$, $\Delta_t=(\mathit{\Sigma}_t-\max_{\tau\in 1:t-1}\mathit{\Sigma}_{\tau})_+$ and $\mathit{\Sigma}_t=\big(\sum_{\tau=1}^{t}\delta_{\tau}\big)_+$ guarantees that 
\begin{equation*}
\begin{aligned}
\textrm{Dual-Gap}\ \{(u_t,v_t)\}_{t=1:T}\leqslant O\left(\min\bigg\{\sum_{t=1}^{T}\rho\left(f_{t}, f_{t-1}\right), P+\sqrt{(1+P)T}\bigg\}\right),
\end{aligned}
\end{equation*}
where $\rho\left(f_{t}, f_{t-1}\right)=\max_{x\in X,\,y\in Y}\left\lvert f_t(x,y)-f_{t-1}(x,y)\right\rvert$ measures the distance between $f_t$ and $f_{t-1}$, and the constant $\epsilon > 0$ prevents $\eta_1$ and $\eta_2$ from being infinite. 
\end{theorem}


\begin{algorithm}[tb]
\caption{IOMDA with Adaptive Learning Rates Guaranteed by Dual-Gap}
\label{alg:IOMDA-DP}
\begin{algorithmic}[1]
\STATE {\bfseries Require:} $1$-strongly convex regularizers $\phi\colon\mathscr{X}\rightarrow\mathbb{R}$ and $\psi\colon\mathscr{Y}\rightarrow\mathbb{R}$ satisfying the assumptions of \cref{lem:IOMDA}, and $\{f_t\}_{t\geqslant1}$ satisfying \cref{ass:gradient-bounded}.
\STATE {\bfseries Initialize:} $P$, $t=0$, $\epsilon>0$, $D=\max\{D_\phi, D_\psi\}$, $L=\max\{L_\phi, L_\psi\}$, $C=2D\left(D+\sqrt{2}L\right)$. 
\REPEAT
\STATE $t\gets t+1$. 
\STATE Solve $\eta_t$ by $(C+L P)/\eta_t=\epsilon + \sum_{\tau=1}^{t-2}\Delta_{\tau}$. 
\STATE Output $\left(x_t,y_t\right)$, and then observe $f_t$. 
\STATE $F_t(x,y)=f_t(x,y)+\frac{1}{\eta_t}B_{\phi}\big(x, x_t^\phi\big)-\frac{1}{\eta_t}B_{\psi}\big(y, y_t^\psi\big)$. 
\STATE Update $x_{t+1}=\arg\min_{x\in X}\max_{y\in Y}F_t(x,y)$. 
\STATE Update $y_{t+1}=\arg\max_{y\in Y}\min_{x\in X}F_t(x,y)$. 
\STATE Calculate $\delta_{t-1}$ according to \cref{eq:delta}. \STATE Get $\Delta_{t-1}$ following the setting of \cref{thm:IOMDA}. 
\STATE Observe the competitor $(u_t,v_t)$. 
\STATE $P_t\gets\sum_{t=1}^T \big(\lVert u_t-u_{t-1}\rVert+\lVert v_t-v_{t-1}\rVert\big)$. 
\UNTIL{$P < P_t$}
\end{algorithmic}
\end{algorithm}

For the case where the path-length $P_T$ can be observed on the fly, we compile our method into \cref{alg:IOMDA-DP}. 

Now we conclude this subsection with the following remarks. 

\begin{remark}
The temporal variability part $\sum_{t=1}^{T}\rho\left(f_{t}, f_{t-1}\right)$ acts to restrain the expansion of the duality gap in environments that exhibit asymptotic stability. 
In particular, in scenarios where the sequence of payoff functions does not change over time, IOMDA enjoys a constant duality gap bound. 
\end{remark}

\begin{remark}
Although \cref{lem:IOMDA,thm:IOMDA} bear resemblance to Lemma~5.1 and Theorem~5.1 of \cite{Campolongo2021closer} respectively, they don't rest on mere application. 
Here are the reasons: 
1) The temporal variability part exhibits the necessity of using the \emph{joint update} depicted in \cref{eq:IOMDA}. 
2) The incorporation of adaptive learning rates requires a relaxation of the upper bound. 
3) The accumulation of $\Delta_t$ requires a one-step delay, ensuring that the learning rate remains independent of future information. 
Proof details can be located in \cref{sec:analysis}.
\end{remark}

\begin{remark}
Consider a scenario where the path-length $P_T$ can be observed on the fly, that is, the competitor can be observed simultaneously as the payoff function is revealed, such as OSP problems with computable saddle points. 
In this setting, we preset an upper limit $P$ for $P_T$ in \cref{alg:IOMDA-DP}, and remove this limitation by using the doubling trick~\citep{schapire1995gambling}. 
Since the doubling trick introduces logarithmic factors at most, the upper bound of Dual-Gap can be succinctly expressed as $\widetilde{O}\big(\min\big\{\sum_{t=1}^{T}\rho\left(f_{t}, f_{t-1}\right), \sqrt{(1+P_T)T}\big\}\big)$ by using the big $\widetilde{O}$ notation to hide poly-logarithmic factors. 
\end{remark}

\begin{remark}
Typically, IOMDA lacks a universal solution and necessitates a case-by-case computation tailored to the specific nature of the payoff functions. 
In essence, each iteration of IOMDA involves addressing a saddle point problem.
In certain instances, IOMDA can be expressed in a closed form, such as when the regularizers $\phi$ and $\psi$ are both square norms, and $f_t$ is bilinear or quadratic. 
When a closed form solution is unavailable, numerical techniques can be employed for efficient approximation \citep{abernethy2018faster,carmon2020coordinate,jin2022sharper}.  
It's noteworthy that when considering a numerical saddle-point solver, additional assumptions of strong convexity and smoothness are typically requisite for payoff functions. 
\end{remark}


\subsection{Optimistic Implicit Online Mirror Descent-Ascent}
\label{sec:OptIOMDA}

\cref{sec:IOMDA} shows that IOMDA achieves the duality gap of $O\big(\min\{\sum_{t=1}^{T}\rho\left(f_{t}, f_{t-1}\right), P+\sqrt{(1+P)T}\}\big)$. 
We expect the worst part to enjoy an $O\big(\sqrt{(1+P)T}\big)$ bound. 
Perhaps improved analysis will do the trick, but we prefer to use optimistic algorithms because of additional benefits of predictors. 
Consider the following optimistic update: 
\begin{equation}
\label{eq:OptIOMDA}
\begin{aligned}
H_t(x,y)&=h_t(x,y)+\frac{1}{\eta_t}B_{\phi}\big(x, \widetilde{x}_t^\phi\big)-\frac{1}{\eta_t}B_{\psi}\big(y, \widetilde{y}_t^\psi\big), \\
(x_{t},y_{t})&=\text{ the saddle point of }H_t\text{ in }X\times Y, \\
\widetilde{x}_{t+1}&=\arg\min_{x\in X} f_t(x,y_t)+\frac{1}{\eta_t}B_{\phi}\big(x, \widetilde{x}_{t}^\phi\big), \\
\widetilde{y}_{t+1}&=\arg\max_{y\in Y} f_t(x_t,y)-\frac{1}{\eta_t}B_{\psi}\big(y, \widetilde{y}_{t}^\psi\big),
\end{aligned}
\end{equation}
where $h_{t}$ is an arbitrary prediction of $f_t$, $\phi\colon\mathscr{X}\rightarrow\mathbb{R}$ and $\psi\colon\mathscr{Y}\rightarrow\mathbb{R}$ are all $1$-strongly convex functions, and $\eta_t>0$ represents the learning rate. 
We call \cref{eq:OptIOMDA} the Optimistic IOMDA~(OptIOMDA). 
The following theorem provides the duality gap guarantee for OptIOMDA. 

\begin{theorem}[Dual-Gap bound for OptIOMDA]
\label{lem:OptIOMDA}
Under the assumptions of \cref{lem:IOMDA}, and let $h_{t}$ be an arbitrary predictor. 
OptIOMDA with non-increasing learning rates guarantees that 
\begin{equation*}
\begin{aligned}
\textrm{Dual-Gap}\ \{(u_t,v_t)\}_{t=1:T}
\leqslant \frac{2D^2+L P_T}{\eta_T} + \sum_{t=1}^{T} \delta_t,
\end{aligned}
\end{equation*}
where $D=\max\{D_\phi, D_\psi\}$, $L=\max\{L_\phi, L_\psi\}$, $P_T= \sum_{t=1}^T\big( \lVert u_t-u_{t-1}\rVert+\lVert v_t-v_{t-1}\rVert\big)$, and  
\begin{equation}
\label{eq:ddelta}
\begin{aligned}
0\leqslant\delta_t=\ &f_t(x_t,\widetilde{y}_{t+1})-h_t(x_t,\widetilde{y}_{t+1})+h_t(\widetilde{x}_{t+1},y_t)-f_t(\widetilde{x}_{t+1}, y_t) \\
&-\big[B_{\psi}\big(\widetilde{y}_{t+1}, y_t^\psi\big)+B_{\phi}\big(\widetilde{x}_{t+1}, x_t^\phi\big)\big] / \eta_t.
\end{aligned}
\end{equation}
\end{theorem}


Note that $\delta_t\geqslant 0$, so one can use adaptive trick adopted in \citet[][Theorem~5.1]{Campolongo2021closer} to set learning rates. 

\begin{theorem}[Dual-Gap bound for OptIOMDA with Adaptive Learning Rates]
\label{thm:OptIOMDA}
Under the assumptions of \cref{lem:OptIOMDA} and let $f_t$ and $h_t$ satisfy \cref{ass:gradient-bounded}, let $P$ be a preset upper bound of $P_T$. 
OptIOMDA with $(2D^2+L P)/\eta_t=\epsilon + \sum_{\tau=1}^{t-1}\delta_{\tau}$ incurs 
\begin{equation*}
\begin{aligned}
\textrm{Dual-Gap}\ \{(u_t,v_t)\}_{t=1:T}\leqslant O\left(\min\bigg\{\sum_{t=1}^{T}\rho\left(f_{t}, h_{t}\right), \sqrt{(1+P)T}\bigg\}\right),
\end{aligned}
\end{equation*}
where $\epsilon>0$ is a constant to prevent $\eta_1$ from being infinite. 
\end{theorem}

\begin{algorithm}[tb]
\caption{OptIOMDA with Adaptive Learning Rates Guaranteed by Dual-Gap}
\label{alg:OptIOMDA-DP}
\begin{algorithmic}[1]
\STATE {\bfseries Require:} $1$-strongly convex regularizers $\phi\colon\mathscr{X}\rightarrow\mathbb{R}$ and $\psi\colon\mathscr{Y}\rightarrow\mathbb{R}$ satisfying the assumptions of \cref{lem:OptIOMDA}, $\{f_t\}_{t\geqslant1}$ and predictor sequence $\{h_t\}_{t\geqslant1}$ satisfying \cref{ass:gradient-bounded}. 
\STATE {\bfseries Initialize:} $P$, $t=0$, $\epsilon>0$, $D=\max\{D_\phi, D_\psi\}$, $L=\max\{L_\phi, L_\psi\}$. 
\REPEAT
\STATE $t\gets t+1$. 
\STATE Solve $\eta_t$ by $(2D^2+L P)/\eta_t=\epsilon + \sum_{\tau=1}^{t-1}\delta_{\tau}$. 
\STATE Receive $h_t$, and let $H_t(x,y)=h_t(x,y)+\frac{1}{\eta_t}B_{\phi}\big(x, \widetilde{x}_{t}^\phi\big)-\frac{1}{\eta_t}B_{\psi}\big(y, \widetilde{y}_{t}^\psi\big)$. 
\STATE Update $x_{t}=\arg\min_{x\in X}\max_{y\in Y}H_t(x,y)$. 
\STATE Update $y_{t}=\arg\max_{y\in Y}\min_{x\in X}H_t(x,y)$. 
\STATE Output $\left(x_t,y_t\right)$, and then observe $f_t$. 
\STATE Update $\widetilde{x}_{t+1}=\arg\min_{x\in X}f_t(x,y_t)+\frac{1}{\eta_t}B_{\phi}\big(x, \widetilde{x}_{t}^\phi\big)$. 
\STATE Update $\widetilde{y}_{t+1}=\arg\max_{y\in Y}f_t(x_t,y)-\frac{1}{\eta_t}B_{\psi}\big(y, \widetilde{y}_{t}^\psi\big)$. 
\STATE Calculate $\delta_t$ according to \cref{eq:ddelta}. 
\STATE Observe the competitor $(u_t,v_t)$. 
\STATE $P_t\gets\sum_{t=1}^T \big(\lVert u_t-u_{t-1}\rVert+\lVert v_t-v_{t-1}\rVert\big)$. 
\UNTIL{$P < P_t$}
\end{algorithmic}
\end{algorithm}

For the case where the path-length $P_T$ can be observed on the fly, we compile OptIOMDA with adaptive learning rates in \cref{alg:OptIOMDA-DP}. 
The remarks below are useful for further understanding of our method. 

\begin{remark}
Recall the optimistic variant of IOMD~(OptIOMD): 
\begin{equation*}
\begin{aligned}
x_{t}=\arg\min_{x\in X}h_t(x)+B_{\phi}\big(x, \widetilde{x}_t^\phi\big)/\eta_t, \quad
\widetilde{x}_{t+1}=\arg\min_{x\in X}\ell_t(x)+B_{\phi}\big(x, \widetilde{x}_t^\phi\big)/\eta_t,
\end{aligned}
\end{equation*}
where $\ell_t$ is the convex loss and $h_{t}$ is an arbitrary prediction of $\ell_t$. 
Given the modification from IOMD~(Equation~\ref{eq:IOMD}) to OptIOMD, one might wonder why not adopt the following update: 
\begin{subequations}
\begin{align}
(x_{t},y_{t})&=\text{ the saddle point of }h_t(x,y)+B_{\phi}\big(x, \widetilde{x}_t^\phi\big)/\eta_t-B_{\psi}\big(y, \widetilde{y}_t^\psi\big)/\eta_t
, \label{eq:update-1}\\
(\widetilde{x}_{t+1},\widetilde{y}_{t+1})&=\text{ the saddle point of }f_t(x,y)+B_{\phi}\big(x, \widetilde{x}_t^\phi\big)/\eta_t-B_{\psi}\big(y, \widetilde{y}_t^\psi\big)/\eta_t
.\label{eq:update-2}
\end{align}
\end{subequations}
We claim that the above update causes difficulties in the Dual-Gap decomposition. 
Indeed, applying first-order optimality condition on \cref{eq:update-1,eq:update-2} requires the introduction of the following two identity transformations, respectively: 
\begin{equation*}
\begin{aligned}
f_t(\widetilde{x}_{t+1}, y_t)-f_t(x_t,\widetilde{y}_{t+1})
&=f_t(\widetilde{x}_{t+1}, y_t)-f_t(\widetilde{x}_{t+1}, \widetilde{y}_{t+1})+f_t(\widetilde{x}_{t+1}, \widetilde{y}_{t+1})-f_t(x_t,\widetilde{y}_{t+1}),\\
h_t(x_t,\widetilde{y}_{t+1})-h_t(\widetilde{x}_{t+1},y_t)
&=h_t(x_t,\widetilde{y}_{t+1})-h_t(x_t,y_t)+h_t(x_t,y_t)-h_t(\widetilde{x}_{t+1},y_t).
\end{aligned}
\end{equation*}
However, it is difficult for them to relate to the duality gap. 
The mismatch lies in $f_t(\widetilde{x}_{t+1}, \widetilde{y}_{t+1})$. 
In addition, we encountered a similar problem when analyzing the upper bound of NE-regret, and it turns out that only \cref{eq:OptIOMDA} works. 
\end{remark}

\begin{remark}
OptIOMDA not only guarantees the Dual-Gap bound, but also ensures individual regrets. 
In fact, the result of \cref{lem:OptIOMDA} can be derived by summing the upper bounds of two individual regrets: 
$f_t(x_t,v_t)-f_t(u_t, y_t) = [f_t(x_t,v_t)-f_t(x_t, y_t)]+[f_t(x_t, y_t)-f_t(u_t, y_t)]$, where
\begin{equation*}
\begin{aligned}
f_t(x_t, y_t)-f_t(u_t, y_t)=\ &h_t(\widetilde{x}_{t+1},y_t)-f_t(\widetilde{x}_{t+1}, y_t)+f_t(\widetilde{x}_{t+1}, y_t)-f_t(u_t, y_t) \\
&+h_t(x_t,y_t)-h_t(\widetilde{x}_{t+1},y_t)+f_t(x_t,y_t)-h_t(x_t,y_t),
\end{aligned}
\end{equation*}
and $f_t(x_t,v_t)-f_t(x_t, y_t)$ can be similarly decomposed. 
It is straightforward to derive the upper bounds of two individual regrets from the proof of \cref{lem:OptIOMDA}.
\end{remark}

\begin{remark}
The duality gap of \cref{thm:OptIOMDA} echoes the dynamic regret of \citet[][Theorem~2.17]{scroccaro2022adaptive}. 
The introduction of the predictor sequence yields additional advantages. 
For example, if we let $h_t = f_{t-2}$, it covers both scenarios where $\rho(f_t,f_{t-1})\rightarrow 0$ and $\rho(f_t,f_{t-2})\rightarrow 0$. 
\end{remark}

\begin{remark}
The worst case bound of \cref{thm:OptIOMDA} is optimal. 
In fact, regardless of the strategies adopted by the players, there always exists an adversarial environment and a competitor sequence satisfying $P_T\leqslant P$, ensuring that the Dual-Gap is not lower than $\mathit{\Omega}\big(\sqrt{(1+P)T}\big)$. 
This assertion can be verified by transforming OCCO into a pair of OCOs. 
Let $\mathscr{L}_X(G)=\big\{\ell\text{\,is convex\,}\big|\,\sup_{x\in X}\left\lVert\partial\ell(x)\right\rVert\leqslant G\big\}$ and 
$\mathcal{F}=\big\{f\,\big|\,f(\,\cdot\,,y)\in\mathscr{L}_X(G_X), \forall y\in Y; -f(x,\,\cdot\,)\in\mathscr{L}_Y(G_Y), \forall x\in X\big\}$, 
let $U_T(P)=\big\{u_{1:T}\,\big| \sum_{t=1}^T\lVert u_t-u_{t-1}\rVert\leqslant P\big\}$ and
$S_T(P)=\big\{(u_t,v_t)_{t\in 1:T}\,\big|\,u_{1:T}\in U_T(p),\,v_{1:T}\in U_T(q),\,p+q\leqslant P\big\}$. 
Now we have 
\begin{subequations}
\begin{align}
&\sup_{f_{1:T}\in\mathcal{F}}\left(\max_{(u_t,v_t)_{t\in 1:T}\in S_T(P)}\sum_{t=1}^T \Big(f_t\left(x_t,v_t\right)- f_t\left(u_t,y_t\right)\Big)\right) \notag\\
\geqslant\ &\sup_{\begin{subarray}{c}f_t(x,y)=\alpha_t(x)-\beta_t(y)\in\mathcal{F}\\t\in 1:T\end{subarray}}\Bigg(\max_{\begin{subarray}{c}u_{1:T}\in U_T(p),\,v_{1:T}\in U_T(q)\\p+q=P\end{subarray}}\sum_{t=1}^T \Big(f_t\left(x_t,v_t\right)- f_t\left(u_t,y_t\right)\Big)\Bigg) \notag\\
=\ &\sup_{\alpha_{1:T}\in\mathscr{L}_X(G_X)}\left(\max_{u_{1:T}\in U_T(p)}\sum_{t=1}^T \Big(\alpha_t\left(x_t\right)- \alpha_t\left(u_t\right)\Big)\right) \label{eq:lower-bound-1}\\
&\ +\sup_{\beta_{1:T}\in\mathscr{L}_Y(G_Y)}\left(\max_{v_{1:T}\in U_T(P-p)}\sum_{t=1}^T \Big(\beta_t\left(y_t\right)- \beta_t\left(v_t\right)\Big)\right).\label{eq:lower-bound-2}
\end{align}
\end{subequations}
According to Theorem~2 in \citet{zhang2018adaptive}, the lower bounds of \cref{eq:lower-bound-1,eq:lower-bound-2} are $\mathit{\Omega}\big(\sqrt{(1+p)T}\big)$ and $\mathit{\Omega}\big(\sqrt{(1+P-p)T}\big)$, respectively. 
These results lead to the $\mathit{\Omega}\big(\sqrt{(1+P)T}\big)$ lower bound for the Dual-Gap. 
\end{remark}

\begin{remark}
As with \cref{alg:IOMDA-DP}, one can eliminate the dependence on $P$ by applying the doubling trick to \cref{alg:OptIOMDA-DP}. 
Consequently, the duality gap is $\widetilde{O}\big(\min\big\{\sum_{t=1}^{T}\rho\left(f_{t}, h_{t}\right), \sqrt{(1+P_T)T}\big\}\big)$. 
The rules for updating $\widetilde{x}_{t+1}$ and $\widetilde{y}_{t+1}$ can be calculated numerically \citep{song2018fully}, or converted to closed form when applicable. 
\end{remark}

\begin{remark}
Consider online matrix games, that is, OCCO with bilinear payoff functions. 
Let's denote the bilinear function $f_t$ as matrix $A_t$ and the predictor $h_t$ as matrix $H_t$. 
Then the upper bound of duality gap can be rearranged as follows: 
\begin{equation*}
\begin{aligned}\textstyle
\textrm{Dual-Gap}\ \{(u_t,v_t)\}_{t=1:T}\leqslant O\left(\min\Big\{\sum_{t=1}^{T}\left\lVert A_{t}-H_{t}\right\rVert, \sqrt{(1+P)\sum_{t=1}^{T}\left\lVert A_{t}-H_{t}\right\rVert^2}\Big\}\right).
\end{aligned}
\end{equation*}
\end{remark}

\subsection{Analysis}
\label{sec:analysis}

This subsection supplements the proofs of \cref{lem:IOMDA,thm:IOMDA,lem:OptIOMDA,thm:OptIOMDA}.
The analysis framework follows \citet[][Appendix~A]{Campolongo2021closer}. 

\begin{proof}[Proof of \cref{lem:IOMDA}]
We first perform identity transformation on the instantaneous duality gap: 
\begin{equation*}
\begin{aligned}
f_t(x_t,v_t)-f_t(u_t, y_t)
=\ & f_t(x_t,v_t)-f_t(x_{t+1}, v_{t+1})+f_t(u_{t+1}, y_{t+1})-f_t(u_t, y_t) \\
&+\underbrace{f_t(x_{t+1}, v_{t+1})-f_t(x_{t+1}, y_{t+1})}_{\hypertarget{termi}{\text{term i}}}+
\underbrace{f_t(x_{t+1}, y_{t+1})-f_t(u_{t+1}, y_{t+1})}_{\hypertarget{termii}{\text{term ii}}}. 
\end{aligned}
\end{equation*}
Applying convexity and first-order optimality condition yields the existence of $x_{t+1}^{f_t(\,\cdot\,,y_{t+1})}$ and $y_{t+1}^{-f_t(x_{t+1},\,\cdot\,)}$, such that
\begin{equation*}
\begin{aligned}
\text{term \hyperlink{termi}{i}}&\leqslant\big\langle y_{t+1}^{-f_t(x_{t+1},\,\cdot\,)}, y_{t+1}-v_{t+1}\big\rangle 
\leqslant\big\langle y_t^\psi-y_{t+1}^\psi, y_{t+1}-v_{t+1}\big\rangle/\eta_t \\
&=\underbrace{\big[B_{\psi}\big(v_{t+1}, y_{t}^\psi\big)-B_{\psi}\big(v_{t+1}, y_{t+1}^\psi\big) \big]/\eta_t}_{\eqcolon\Delta_t^y}
-B_{\psi}\big(y_{t+1}, y_{t}^\psi\big)/\eta_t, \\
\text{term \hyperlink{termii}{ii}}&\leqslant\big\langle x_{t+1}^{f_t(\,\cdot\,,y_{t+1})}, x_{t+1}-u_{t+1}\big\rangle 
\leqslant\big\langle x_t^\phi-x_{t+1}^\phi, x_{t+1}-u_{t+1}\big\rangle/\eta_t \\
&=\underbrace{\big[B_{\phi}\big(u_{t+1}, x_{t}^\phi\big)-B_{\phi}\big(u_{t+1}, x_{t+1}^\phi\big) \big]/\eta_t}_{\eqcolon\Delta_t^x}
-B_{\phi}\big(x_{t+1}, x_{t}^\phi\big)/\eta_t.
\end{aligned}
\end{equation*}
Substituting terms~\hyperlink{termi}{i} and~\hyperlink{termii}{ii} into the instantaneous duality gap and summing over time yields
\begin{equation*}
\begin{aligned}
\sum_{t=1}^{T}\Big(f_t(x_t,v_t)-f_t(u_t, y_t)\Big)
\leqslant\sum_{t=1}^{T}\Delta_t^x + \sum_{t=1}^{T}\Delta_t^y
+\sum_{t=1}^{T}\delta_t,
\end{aligned}
\end{equation*}
where 
\begin{equation*}
\begin{aligned}
\sum_{t=1}^{T}\Delta_t^x
&\leqslant \sum_{t=1}^{T}\frac{1}{\eta_t}\left[B_{\phi}\big(u_{t+1}, x_{t}^\phi\big)-B_{\phi}\big(u_{t}, x_{t}^\phi\big)\right] + \frac{B_{\phi}\big(x_{1}^*, x_{1}^\phi\big)}{\eta_0} + \sum_{t=1}^{T}\Big(\frac{1}{\eta_{t}}-\frac{1}{\eta_{t-1}}\Big)B_{\phi}\big(u_{t}, x_{t}^\phi\big) \\
&\leqslant\frac{D_\phi^2}{\eta_T}+\sum_{t=1}^{T}\frac{L_\phi}{\eta_t}\left\lVert u_{t+1}-u_{t}\right\rVert
\leqslant \frac{D_\phi}{\eta_T}\left(D_\phi+\sqrt{2}L_\phi\right)+\sum_{t=1}^{T}\frac{L_\phi}{\eta_t}\left\lVert u_{t}-u_{t-1}\right\rVert
\end{aligned}
\end{equation*}
since $\eta_t$ is non-increasing over time, $B_\phi$ is $L_\phi$-Lipschitz w.r.t. the first variable, and $D_\phi^2$ is the supremum of $B_\phi$. Similarly, 
\begin{equation*}
\begin{aligned}
\sum_{t=1}^{T}\Delta_t^y
\leqslant\frac{D_\psi}{\eta_T}\left(D_\psi+\sqrt{2}L_\psi\right)+\sum_{t=1}^{T}\frac{L_\psi}{\eta_t}\left\lVert v_{t}-v_{t-1}\right\rVert.
\end{aligned}
\end{equation*}
Let $P_T= \sum_{t=1}^T \big(\lVert u_t-u_{t-1}\rVert+\lVert v_t-v_{t-1}\rVert\big)$. 
Therefore, we get 
\begin{equation*}
\textrm{Dual-Gap}\ \{(u_t,v_t)\}_{t=1:T}
=\sum_{t=1}^{T}\Big(f_t(x_t,v_t)-f_t(u_t, y_t)\Big)
\leqslant \frac{2D\left(D+\sqrt{2}L\right)+L P_T}{\eta_T} + \sum_{t=1}^{T}\delta_t. \qedhere
\end{equation*}
\end{proof}

\begin{proof}[Proof of \cref{thm:IOMDA}]
First, we claim that $\sum_{\tau=1}^t \Delta_\tau = \max_{\tau\in 1:t}\mathit{\Sigma}_{\tau}$. 
This claim can be proved by induction. 
It is obvious that $\Delta_1=\mathit{\Sigma}_1$. 
Now we assume the claim holds for $t-1$ and prove it for $t$: 
\begin{equation*}
\begin{aligned}
\sum_{\tau=1}^t \Delta_\tau
=\Delta_t + \sum_{\tau=1}^{t-1} \Delta_\tau 
=\Big(\mathit{\Sigma}_t-\max_{\tau\in 1:t-1}\mathit{\Sigma}_{\tau}\Big)_+ + \max_{\tau\in 1:t-1}\mathit{\Sigma}_{\tau}
= \max_{\tau\in 1:t}\mathit{\Sigma}_{\tau}.
\end{aligned}
\end{equation*}
Applying the prescribed learning rate $\eta_t$ , we relax the bound in \cref{lem:IOMDA} as follows: 
\begin{equation*}
\begin{aligned}
\textrm{Dual-Gap}\ \{(u_t,v_t)\}_{t=1:T}
\leqslant \epsilon+\sum_{t=1}^{T-2} \Delta_t + \max_{t\in 1:T}\mathit{\Sigma}_{t} 
\leqslant \epsilon+2\sum_{t=1}^{T} \Delta_t 
= \epsilon+2\max_{t\in 1:T}\mathit{\Sigma}_{t}. 
\end{aligned}
\end{equation*}
Next, we estimate the upper bound of the r.h.s. of the above inequality in two ways. 
The firse one: 
\begin{equation*}
\begin{aligned}
\max_{t\in 1:T}\mathit{\Sigma}_{t}\leqslant O\left(1\right)+2\sum_{t=1}^{T}\rho\left( f_{t},f_{t-1}\right). 
\end{aligned}
\end{equation*}
Indeed, the monotonicity of $(\,\cdot\,)_+$ implies that 
\begin{subequations}
\begin{align}
\mathit{\Sigma}_{t} 
\leqslant\ &\bigg(\sum_{\tau=1}^{t}\big[f_\tau(x_\tau,v_\tau)-f_\tau(x_{\tau+1}, v_{\tau+1})+f_\tau(u_{\tau+1}, y_{\tau+1})-f_\tau(u_\tau, y_\tau)\big]\bigg)_+ \notag\\
\leqslant\ &\big|f_{0}(x_1,v_1)-f_{0}(u_1,y_1)-f_{t}(x_{t+1}, v_{t+1})+f_{t}(u_{t+1}, y_{t+1})\big| \label{eq:sigma1}\\
&+ \sum_{\tau=1}^{t}\big|f_{\tau}(x_\tau,v_\tau)-f_{\tau-1}(x_{\tau}, v_{\tau})\big|+\sum_{\tau=1}^{t}\big|f_{\tau}(u_\tau,y_\tau)-f_{\tau-1}(u_{\tau}, y_{\tau})\big|, \label{eq:sigma2}
\end{align}
\end{subequations}
where 
\begin{equation*}
\begin{aligned}
\textrm{Eq.~\eqref{eq:sigma1}}
\leqslant\ &\big|f_{0}(x_1,v_1)-f_{0}(x_1,y_1)+f_{0}(x_1,y_1)-f_{0}(u_1,y_1)\big| \\
&+\big|f_{t}(x_{t+1}, v_{t+1})-f_{t}(x_{t+1}, y_{t+1})+f_{t}(x_{t+1}, y_{t+1})-f_{t}(u_{t+1}, y_{t+1})\big| \\
\leqslant\ &
\left\lVert y_1-v_1\right\rVert \max\big\{\left\lVert\partial_y (-f_0)(x_1,y_1)\right\rVert, \left\lVert\partial_y (-f_0)(x_1,v_1)\right\rVert\big\} \\
&+\left\lVert x_1-u_1\right\rVert \max\big\{\left\lVert\partial_x f_0(x_1,y_1)\right\rVert, \left\lVert\partial_x f_0(u_1,y_1)\right\rVert\big\} \\
&+\left\lVert y_{t+1}-v_{t+1}\right\rVert \max\big\{\left\lVert\partial_y (-f_t)(x_{t+1},y_{t+1})\right\rVert, \left\lVert\partial_y (-f_t)(x_{t+1},v_{t+1})\right\rVert\big\} \\
&+\left\lVert x_{t+1}-u_{t+1}\right\rVert \max\big\{\left\lVert\partial_x f_t(x_{t+1},y_{t+1})\right\rVert, \left\lVert\partial_x f_t(u_{t+1},y_{t+1})\right\rVert\big\} \\
\leqslant\ &2\sqrt{2}\left(D_\phi G_X+D_\psi G_Y\right), \\
\textrm{Eq.~\eqref{eq:sigma2}}
\leqslant\ &2\sum_{\tau=1}^{t}\rho\left( f_{\tau},f_{\tau-1}\right). 
\end{aligned}
\end{equation*}
The other one: 
\begin{equation}
\label{eq:dual-gap-bound2}
\begin{aligned}
\sum_{t=1}^{T}\Delta_t\leqslant O(P+\sqrt{(1+P)T}\big). 
\end{aligned}
\end{equation}
The derivation is as follows. 
\begin{subequations}
\begin{align}
\Delta_t
=\ &\Big(\mathit{\Sigma}_t-\max_{\tau\in 1:t-1}\mathit{\Sigma}_{\tau}\Big)_+
\leqslant(\mathit{\Sigma}_t-\mathit{\Sigma}_{t-1})_+
\leqslant (\delta_t)_+ \notag\\
\leqslant\ &\big(f_t(x_t,v_t)-f_t(x_{t+1}, v_{t+1})
-B_{\phi}\big(x_{t+1}, x_{t}^\phi\big)/\eta_t\big)_+ \label{eq:delta-x}\\
&+\big(f_t(u_{t+1}, y_{t+1})-f_t(u_t, y_t)
-B_{\psi}\big(y_{t+1}, y_{t}^\psi\big)/\eta_t\big)_+, \label{eq:delta-y}
\end{align}
\end{subequations}
where the second ``$\leqslant$'' follows from a result which says that given $a,b\in\mathbb{R}$, $((a+b)_+-(a)_+)_+\leqslant(b)_+$ holds since $(a+b)_+\leqslant(a)_+ +(b)_+$. 
For \cref{eq:delta-x,eq:delta-y}, we have that 
\begin{equation*}
\begin{aligned}
\textrm{Eq.~\eqref{eq:delta-x}}
&\leqslant\left(\big| f_t(x_t,v_t)-f_t(x_{t}, v_{t+1})\big|+\big| f_t(x_{t}, v_{t+1})-f_t(x_{t+1}, v_{t+1})\big|-B_{\phi}\big(x_{t+1}, x_{t}^\phi\big)/\eta_t\right)_+ \\
&\leqslant \left(G_Y \left\lVert v_t-v_{t+1}\right\rVert+G_X \left\lVert x_t-x_{t+1}\right\rVert-B_{\phi}\big(x_{t+1}, x_{t}^\phi\big)/\eta_t\right)_+ \\
&\leqslant \Big(G_Y \left\lVert v_t-v_{t+1}\right\rVert+G_X\sqrt{2B_{\phi}\big(x_{t+1}, x_{t}^\phi\big)}-B_{\phi}\big(x_{t+1}, x_{t}^\phi\big)/\eta_t\Big)_+ \\
&\leqslant G_Y \left\lVert v_t-v_{t+1}\right\rVert+\min\big\{\sqrt{2}D_\phi G_X,\,\eta_t G_X^2/2\big\},
\end{aligned}
\end{equation*}
and similarly, 
\begin{equation*}
\begin{aligned}
\textrm{Eq.~\eqref{eq:delta-y}}
\leqslant G_X \left\lVert u_t-u_{t+1}\right\rVert+\min\big\{\sqrt{2}D_\psi G_Y,\,\eta_t G_Y^2/2\big\}.
\end{aligned}
\end{equation*}
Let $\xi_t=\big(\Delta_t-G \big(\left\lVert u_t-u_{t+1}\right\rVert+\left\lVert v_t-v_{t+1}\right\rVert\big)\big)_+$, then 
\begin{equation*}
\begin{aligned}
\xi_t\leqslant\min\big\{2\sqrt{2}D G,\,\eta_t G^2\big\},
\end{aligned}
\end{equation*}
and thus, 
\begin{equation*}
\begin{aligned}
\left(\sum_{t=1}^{T-1}\xi_t\right)^2
&=\sum_{t=1}^{T-1}\xi_t^2+2\sum_{t=1}^{T-1}\xi_t\sum_{\tau=1}^{t-2}\xi_\tau
\leqslant\sum_{t=1}^{T-1}\xi_t^2+2\sum_{t=1}^{T-1}\xi_t\sum_{\tau=1}^{t-2}\Delta_\tau \\
&=\sum_{t=1}^{T-1}\xi_t^2+2\sum_{t=1}^{T-1}\xi_t \left(\frac{C+L P}{\eta_t}-\epsilon\right) 
\leqslant \sum_{t=1}^{T-1}8 D^2 G^2+\sum_{t=1}^{T-1} 2G^2(C+L P),
\end{aligned}
\end{equation*}
where the last ``$\leqslant$'' uses the first and second terms in the minimum of the bound for $\xi_t$ in turn. 
Now we get 
\begin{equation*}
\begin{aligned}
\sum_{t=1}^{T}\Delta_t\leqslant O(P)+\sum_{t=1}^{T}\xi_t\leqslant O(P)+O\big(\sqrt{(1+P)T}\big).
\end{aligned}
\end{equation*}
So \cref{eq:dual-gap-bound2} holds. 
In conclusion, 
\begin{equation*}
\textrm{Dual-Gap}\ \{(u_t,v_t)\}_{t=1:T}
\leqslant O\left(\min\bigg\{\sum_{t=1}^{T}\rho\left(f_{t}, f_{t-1}\right), P+\sqrt{(1+P)T}\bigg\}\right). \qedhere
\end{equation*}
\end{proof}

\begin{proof}[Proof of \cref{lem:OptIOMDA}]
We first perform identity transformation on the instantaneous duality gap: 
\begin{equation*}
\begin{aligned}
f_t(x_t,v_t)&-f_t(u_t, y_t) =
f_t(x_t,\widetilde{y}_{t+1})-h_t(x_t,\widetilde{y}_{t+1})+h_t(\widetilde{x}_{t+1},y_t)-f_t(\widetilde{x}_{t+1}, y_t) \\
&+\underbrace{f_t(\widetilde{x}_{t+1}, y_t)-f_t(u_t, y_t)}_{\hypertarget{term1}{\text{term 1}}}
+\underbrace{f_t(x_t, v_t)-f_t(x_t,\widetilde{y}_{t+1})}_{\hypertarget{term2}{\text{term 2}}}
+\underbrace{h_t(x_t,\widetilde{y}_{t+1})-h_t(\widetilde{x}_{t+1},y_t)}_{\hypertarget{term3}{\text{term 3}}},
\end{aligned}
\end{equation*}
Applying convexity and first-order optimality condition yields the existence of $\widetilde{x}_{t+1}^{f_t(\,\cdot\,, y_t)}$, $\widetilde{y}_{t+1}^{-f_t(x_t,\,\cdot\,)}$, $x_t^{h_t(\,\cdot\,,y_t)}$ and $y_t^{-h_t(x_t,\,\cdot\,)}$, such that 
\begin{equation*}
\begin{aligned}
\text{term \hyperlink{term1}{1}}&\leqslant\big\langle\widetilde{x}_{t+1}^{f_t(\,\cdot\,, y_t)}, \widetilde{x}_{t+1}-u_t\big\rangle 
\leqslant\big\langle \widetilde{x}_{t}^\phi-\widetilde{x}_{t+1}^\phi,\widetilde{x}_{t+1}-u_t\big\rangle / \eta_t \\
&=\underbrace{\big[B_{\phi}\big(u_t, \widetilde{x}_{t}^\phi\big)-B_{\phi}\big(u_t, \widetilde{x}_{t+1}^\phi\big)\big] / \eta_t}_{\eqcolon\Delta_t^x}-B_{\phi}\big(\widetilde{x}_{t+1}, \widetilde{x}_{t}^\phi\big) / \eta_t, \\
\text{term \hyperlink{term2}{2}}&\leqslant\big\langle\widetilde{y}_{t+1}^{-f_t(x_t,\,\cdot\,)}, \widetilde{y}_{t+1}-v_t\big\rangle 
\leqslant\big\langle \widetilde{y}_{t}^\psi-\widetilde{y}_{t+1}^\psi,\widetilde{y}_{t+1}-v_t\big\rangle / \eta_t \\
&=\underbrace{\big[B_{\psi}\big(v_t, \widetilde{y}_{t}^\psi\big)-B_{\psi}\big(v_t, \widetilde{y}_{t+1}^\psi\big)\big] / \eta_t}_{\eqcolon\Delta_t^y}-B_{\psi}\big(\widetilde{y}_{t+1}, \widetilde{y}_{t}^\psi\big) / \eta_t, \\
\text{term \hyperlink{term3}{3}}
&=h_t(x_t,\widetilde{y}_{t+1})-h_t(x_t,y_t)+h_t(x_t,y_t)-h_t(\widetilde{x}_{t+1},y_t) \\
&\leqslant\big\langle y_t^{-h_t(x_t,\,\cdot\,)}, y_t-\widetilde{y}_{t+1}\big\rangle+\big\langle x_t^{h_t(\,\cdot\,,y_t)}, x_t-\widetilde{x}_{t+1}\big\rangle \\
&\leqslant\big\langle \widetilde{y}_{t}^\psi-y_{t}^\psi,y_t-\widetilde{y}_{t+1}\big\rangle/ \eta_t + \big\langle \widetilde{x}_{t}^\phi-x_{t}^\phi,x_t-\widetilde{x}_{t+1}\big\rangle/ \eta_t \\
&=\big[B_{\psi}\big(\widetilde{y}_{t+1}, \widetilde{y}_t^\psi\big)-B_{\psi}\big(\widetilde{y}_{t+1}, y_t^\psi\big)-B_{\psi}\big(y_t, \widetilde{y}_t^\psi\big)\big]/ \eta_t \\
&\hspace*{1.3em}+ \big[B_{\phi}\big(\widetilde{x}_{t+1}, \widetilde{x}_t^\phi\big)-B_{\phi}\big(\widetilde{x}_{t+1}, x_t^\phi\big)-B_{\phi}\big(x_t, \widetilde{x}_t^\phi\big)\big]/ \eta_t,
\end{aligned}
\end{equation*}
Substituting terms~\hyperlink{term1}{1}, \hyperlink{term2}{2} and~\hyperlink{term3}{3} into the instantaneous duality gap and summing over time yields
\begin{equation*}
\begin{aligned}
\sum_{t=1}^{T}\Big(f_t(x_t,v_t)-f_t(u_t, y_t)\Big)
\leqslant\sum_{t=1}^{T}\Delta_t^x + \sum_{t=1}^{T}\Delta_t^y
+\sum_{t=1}^{T}\delta_t, 
\end{aligned}
\end{equation*}
where 
\begin{equation*}
\begin{aligned}
\sum_{t=1}^{T}\Delta_t^x
&\leqslant\sum_{t=1}^{T}\frac{1}{\eta_t}\left[B_{\phi}\big(u_t, \widetilde{x}_{t}^\phi\big)-B_{\phi}\big(u_{t-1}, \widetilde{x}_{t}^\phi\big)\right] 
+\frac{B_{\phi}\big(u_{0}, \widetilde{x}_{1}^\phi\big)}{\eta_0}\\
&\hspace*{1.2em}+\sum_{t=1}^{T}\left(\frac{1}{\eta_{t}}-\frac{1}{\eta_{t-1}}\right)B_{\phi}\big(u_{t-1}, \widetilde{x}_{t}^\phi\big) 
\leqslant\frac{D_\phi^2}{\eta_T}+\sum_{t=1}^{T}\frac{L_\phi}{\eta_t}\left\lVert u_t-u_{t-1}\right\rVert
\end{aligned}
\end{equation*}
since $\eta_t$ is non-increasing over time, $B_\phi$ is $L_\phi$-Lipschitz w.r.t. the first variable, and $D_\phi^2$ is the supremum of $B_\phi$. Similarly, 
\begin{equation*}
\begin{aligned}
\sum_{t=1}^{T}\Delta_t^y
\leqslant\frac{D_\psi^2}{\eta_T}+\sum_{t=1}^{T}\frac{L_\psi}{\eta_t}\left\lVert v_t-v_{t-1}\right\rVert.
\end{aligned}
\end{equation*}
Now we get 
\begin{equation*}
\begin{aligned}
\textrm{Dual-Gap}\ \{(u_t,v_t)\}_{t=1:T}
\leqslant \frac{2D^2+L P_T}{\eta_T} + \sum_{t=1}^{T} \delta_t.
\end{aligned}
\end{equation*}
To verify $\delta_t\geqslant 0$, it suffices to combine the optimality of $\widetilde{x}_{t+1}$, $\widetilde{y}_{t+1}$, and the inequality related to \text{term~\hyperlink{term3}{3}}. 
Indeed, according to the optimality of $\widetilde{x}_{t+1}$ and $\widetilde{y}_{t+1}$, we get 
\begin{equation*}
\begin{aligned}
f_t(x_t,y_t)+B_{\phi}\big(x_t, \widetilde{x}_{t}^\phi\big)/\eta_t\geqslant f_t(\widetilde{x}_{t+1},y_t)+B_{\phi}\big(\widetilde{x}_{t+1}, \widetilde{x}_{t}^\phi\big)/\eta_t, \\
f_t(x_t,y_t)-B_{\psi}\big(y_t, \widetilde{y}_{t}^\psi\big)/\eta_t\leqslant f_t(x_t, \widetilde{y}_{t+1})-B_{\psi}\big(\widetilde{y}_{t+1}, \widetilde{y}_{t}^\psi\big)/\eta_t.
\end{aligned}
\end{equation*}
The inequality related to term~\hyperlink{term3}{3} says 
\begin{equation*}
\begin{aligned}
-h_t(x_t,\widetilde{y}_{t+1})+h_t(\widetilde{x}_{t+1},y_t)
\geqslant& -\big[B_{\psi}\big(\widetilde{y}_{t+1}, \widetilde{y}_t^\psi\big)-B_{\psi}\big(\widetilde{y}_{t+1}, y_t^\psi\big)-B_{\psi}\big(y_t, \widetilde{y}_t^\psi\big)\big] / \eta_t \\
&-\big[B_{\phi}\big(\widetilde{x}_{t+1}, \widetilde{x}_t^\phi\big)-B_{\phi}\big(\widetilde{x}_{t+1}, x_t^\phi\big)-B_{\phi}\big(x_t, \widetilde{x}_t^\phi\big)\big]/ \eta_t.
\end{aligned}
\end{equation*}
Combining the above three inequalities immediately yields $\delta_t\geqslant 0$. 
\end{proof}

\begin{proof}[Proof of \cref{thm:OptIOMDA}]
The prescribed learning rate guarantees that 
\begin{equation*}
\begin{aligned}
\textrm{Dual-Gap}\ \{(u_t,v_t)\}_{t=1:T}
\leqslant \epsilon+2\sum_{t=1}^{T} \delta_t. 
\end{aligned}
\end{equation*}
We provide two upper bounds for $\delta_t$. 
The first one: $\delta_t\leqslant 2\rho\left(f_t, h_t\right)$. 
The second one: $\delta_t\leqslant 2\min\left\{\sqrt{2}D G,\,\eta_t G^2\right\}$. Indeed, 
\begin{equation*}
\begin{aligned}
\delta_t=\ &f_t(x_t,\widetilde{y}_{t+1})-f_t(x_t, y_t)+h_t(x_t, y_t)-h_t(x_t,\widetilde{y}_{t+1})-B_{\psi}\big(\widetilde{y}_{t+1}, y_t^\psi\big)/\eta_t \\
&+f_t(x_t, y_t)-f_t(\widetilde{x}_{t+1}, y_t)+h_t(\widetilde{x}_{t+1},y_t)-h_t(x_t, y_t)-B_{\phi}\big(\widetilde{x}_{t+1}, x_t^\phi\big)/\eta_t \\
\leqslant\ &\big\langle \partial_y (-f_t)(x_t, y_t)-\partial_y (-h_t)(x_t,\widetilde{y}_{t+1}), y_t-\widetilde{y}_{t+1}\big\rangle-B_{\psi}\big(\widetilde{y}_{t+1}, y_t^\psi\big)/\eta_t \\
&+\big\langle \partial_x f_t(x_t, y_t)-\partial_x h_t(\widetilde{x}_{t+1},y_t), x_t-\widetilde{x}_{t+1}\big\rangle-B_{\phi}\big(\widetilde{x}_{t+1}, x_t^\phi\big)/\eta_t \\
\leqslant\ &\big\lVert \partial_y (-f_t)(x_t, y_t)-\partial_y (-h_t)(x_t,\widetilde{y}_{t+1})\big\rVert\sqrt{2 B_{\psi}\big(\widetilde{y}_{t+1}, y_t^\psi\big)}-B_{\psi}\big(\widetilde{y}_{t+1}, y_t^\psi\big)/\eta_t \\
&+\big\lVert \partial_x f_t(x_t, y_t)-\partial_x h_t(\widetilde{x}_{t+1}, y_t)\big\rVert\sqrt{2 B_{\phi}\big(\widetilde{x}_{t+1}, x_t^\phi\big)}-B_{\phi}\big(\widetilde{x}_{t+1}, x_t^\phi\big)/\eta_t \\
\leqslant\ & 2\min\left\{\sqrt{2}\left(D_\phi G_X+D_\psi G_Y\right),\,\eta_t \left(G_X^2+G_Y^2\right)\right\}
\leqslant 4\min\left\{\sqrt{2}DG,\,\eta_t G^2\right\},
\end{aligned}
\end{equation*}
Sum over time for the two cases separately, we get 
\begin{equation*}
\begin{aligned}
\sum_{t=1}^{T}\delta_t\leqslant 2\sum_{t=1}^{T}\rho\left(\ell_t, h_t\right),
\end{aligned}
\end{equation*}
and 
\begin{equation*}
\begin{aligned}
\left(\sum_{t=1}^{T}\delta_t\right)^2
&=\sum_{t=1}^{T}\delta_t^2+2\sum_{t=1}^{T}\delta_t\sum_{\tau=1}^{t-1}\delta_\tau=\sum_{t=1}^{T}\delta_t^2+2\sum_{t=1}^{T}\delta_t\left(\frac{2D^2+L P}{\eta_t}-\epsilon\right) \\
&\leqslant \sum_{t=1}^{T}32 D^2 G^2+\sum_{t=1}^{T}8G^2(2D^2+L P)
= 8G^2(6D^2+L P)T. 
\end{aligned}
\end{equation*}
In conclusion, 
\begin{equation*}
\textrm{Dual-Gap}\ \{(u_t,v_t)\}_{t=1:T}
\leqslant \epsilon+4\min\left\{\sum_{t=1}^{T}\rho\left(f_t, h_t\right),\,G\sqrt{2(6D^2+L P)T}\right\}. 
\qedhere
\end{equation*}
\end{proof}



\section{Experiments}

This section conducts experiments to validate the effectiveness of our proposed algorithms. 
We will begin by describing the experimental setup and then proceed to present the results. 

\subsection{Setup}
\label{sec:setup}

\begin{figure}[htbp]
\centering
{\hfill
\begin{subfigure}[H]{0.3\textwidth}
\caption{Case \hyperlink{caseI}{I}}
\begin{tikzpicture}
\footnotesize
\begin{axis}[width=1.3\textwidth,axis lines=middle,axis equal,xmin=-2,xmax=2,ymin=-2.1,ymax=2.3]
\addplot+[color=red!65!orange,mark options={draw=none},only marks,mark=*,mark size=0.5pt]
table[x=x,y=y,col sep=comma] {Env11.csv};
\end{axis}
\end{tikzpicture}
\label{fig:envir1}
\end{subfigure}
\hfill
\begin{subfigure}[H]{0.3\textwidth}
\caption{Case \hyperlink{caseII}{II}}
\begin{tikzpicture}
\footnotesize
\begin{axis}[width=1.3\textwidth,axis lines=middle,axis equal,xmin=-2,xmax=2,ymin=-2.1,ymax=2.3]
\addplot+[color=red!65!orange,mark options={draw=none},only marks,mark=*,mark size=0.5pt]
table[x=x,y=y,col sep=comma] {Env21.csv};
\addplot+[color=red!65!orange,mark options={draw=none},only marks,mark=*,mark size=0.5pt]
table[x=x,y=y,col sep=comma] {Env22.csv};
\end{axis}
\end{tikzpicture}
\label{fig:envir2}
\end{subfigure}
\hfill
\begin{subfigure}[H]{0.3\textwidth}
\caption{Case \hyperlink{caseIII}{III}}
\begin{tikzpicture}
\footnotesize
\begin{axis}[width=1.3\textwidth,axis lines=middle,axis equal,xmin=-2,xmax=2,ymin=-2.1,ymax=2.3]
\addplot+[color=red!65!orange,mark options={draw=none},only marks,mark=*,mark size=0.5pt]
table[x=x,y=y,col sep=comma] {Env31.csv};
\addplot+[color=red!65!orange,mark options={draw=none},only marks,mark=*,mark size=0.5pt]
table[x=x,y=y,col sep=comma] {Env32.csv};
\addplot+[color=red!65!orange,mark options={draw=none},only marks,mark=*,mark size=0.5pt]
table[x=x,y=y,col sep=comma] {Env33.csv};
\end{axis}
\end{tikzpicture}
\label{fig:envir3}
\end{subfigure}
\hfill}
\caption{Trajectories of Saddle Points. }
\label{fig:Trajectories}
\end{figure}
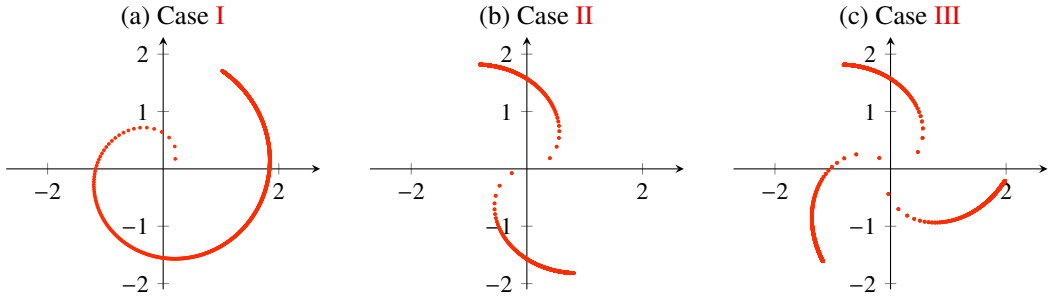

Consider the following synthesis problem: at round $t$, players~1 and~2 jointly select a pair of strategies $(x_t, y_t)$, and then the environment feeds back a convex-concave payoff function $f_t$:
\begin{equation*}
\begin{aligned}
f_t\left(x,y\right)=\frac{1}{2}\left(x-x_t^*\right)^2-\frac{1}{2}\left(y-y_t^*\right)^2+\left(x-x_t^*\right)\left(y-y_t^*\right), 
\end{aligned}
\end{equation*}
$(x_t^*,y_t^*)\in X\times Y$ is the saddle point of $f_t$. 
Let $X=\left[-4, 4\right]$, $Y=\left[-4, 4\right]$. 
It is evident that the requirement of \cref{ass:gradient-bounded} is satisfied. 
To determine the payoff function $f_t$, it suffices to fix the saddle point $(x_t^*,y_t^*)$. 
We set up four cases, which are listed in \cref{tab:cases}. 
Cases~\hyperlink{caseI}{I}, \hyperlink{caseII}{II}, and~\hyperlink{caseIII}{III} correspond to scenarios where $\rho(f_t,f_{t-1})\rightarrow 0$, $\rho(f_t,f_{t-2})\rightarrow 0$, and $\rho(f_t,f_{t-3})\rightarrow 0$, respectively. 
Therefore, we select saddle points as competitors, effectively transforming them into OSP problems. 
The trajectories of these saddle points are displayed in \cref{fig:Trajectories}.
Case~\hyperlink{caseIV}{IV} corresponds to a hostile environment. 
In Case~\hyperlink{caseI}{I}, the saddle points gradually reduce movement speed over time, resulting in sublinear growth of the path length $P_T$. 
In Cases~\hyperlink{caseII}{II} and~\hyperlink{caseIII}{III}, the saddle points cyclically jump between two branches and three branches, respectively, over time. 
It is worth noting that the path length for these two cases exhibits slightly superlinear growth. 
However, due to the significantly slow growth of the log-log function $z_2$, within the limited range of our experiments, we can still approximate the path length's growth as linear. 
For Case~\hyperlink{caseIV}{IV}, to avoid a linear growth of the Dual-Gap, we use saddle points scaled by the contraction factor $1/t$ as competitors, ensuring sublinear growth of the path length $P_T$. 

\begin{table}[tb]
\centering
\caption{Four Settings. }
\label{tab:cases}
\small
\begin{tabular}{clcll}
\toprule
Case & Saddle Point & $(u_t, v_t)$ & $P_T$ & Remark \\
\midrule
\hypertarget{caseI}{I} & $\displaystyle\begin{aligned}
x_t^* &= \mathrm{Re} \big[z_2(t)\mathrm{e}^{i z_1(t)}\big] \\
y_t^* &= \mathrm{Im} \big[z_2(t)\mathrm{e}^{i z_1(t)}\big]
\end{aligned}$
& $(x_t^*, y_t^*)$ & 
Sublinear & $\displaystyle\begin{aligned}&\text{OSP problem with} \\
&\rho(f_t,f_{t-1})\rightarrow 0
\end{aligned}$ \\
\midrule
\hypertarget{caseII}{II} & $\displaystyle\begin{aligned}
x_t^* &= \mathrm{Re} \big[z_2(t)\mathrm{e}^{i\pi t+i z_2(t)}\big] \\
y_t^* &= \mathrm{Im} \big[z_2(t)\mathrm{e}^{i\pi t+i z_2(t)}\big]
\end{aligned}$
& $(x_t^*, y_t^*)$
& $\displaystyle\begin{aligned}&\text{Weak } \\
&\text{Superlinear}
\end{aligned}$ & $\displaystyle\begin{aligned}&\text{OSP problem with} \\
&\rho(f_t,f_{t-2})\rightarrow 0
\end{aligned}$ \\
\midrule
\hypertarget{caseIII}{III} & $\displaystyle\begin{aligned}
x_t^* &= \mathrm{Re} \big[z_2(t)\mathrm{e}^{i\frac{2\pi}{3} t+i z_2(t)}\big] \\
y_t^* &= \mathrm{Im} \big[z_2(t)\mathrm{e}^{i\frac{2\pi}{3} t+i z_2(t)}\big]
\end{aligned}$
& $(x_t^*, y_t^*)$ 
& $\displaystyle\begin{aligned}&\text{Weak } \\
&\text{Superlinear}
\end{aligned}$ & $\displaystyle\begin{aligned}&\text{OSP problem with} \\
&\rho(f_t,f_{t-3})\rightarrow 0
\end{aligned}$ \\
\midrule
\hypertarget{caseIV}{IV} & $\displaystyle\begin{aligned}
x_t^* &=\boldsymbol{1}_{x_t<0}-\boldsymbol{1}_{x_t\geqslant0} \\
y_t^* &=\boldsymbol{1}_{y_t<0}-\boldsymbol{1}_{y_t\geqslant0}
\end{aligned}$ 
& $\displaystyle\frac{1}{t}(x_t^*, y_t^*)$ & Sublinear & Adversarial OCCO \\
\bottomrule
\end{tabular}
\footnotetext{Note: In this table, $z_1(t)=\ln(1+t)$ is a logarithmic growth function of $t$, and $z_2(t)=\ln\ln(\mathrm{e}+t)$ is a log-logarithmic growth function of $t$. $i$ represents the imaginary unit and satisfying the equation $i^{2}=-1$. $\mathrm{Re}$ and $\mathrm{Im}$ represent operators that get real part and imaginary part, respectively. $\boldsymbol{1}_q$ is the $0\,/\,1$ indicator function and satisfying $\boldsymbol{1}_q=1$ iff $q$ is true. }
\end{table}

Now we instantiate IOMDA and OptIOMDA. 
Let $\phi$ and $\psi$ be quadratic functions, i.e., $\phi(x)=x^2/2$, $\psi(y)=y^2/2$, then $B_\phi$ on $X$ and $B_\psi$ on $Y$ are $8$-Lipschitz w.r.t. their first variables and bounded by $32$. 
For OptIOMDA, we configure three types of predictors: 
\begin{itemize}[itemsep=0pt,topsep=0pt]
\item[1.] $h_t = f_{t-1}$.
\item[2.] $h_t = f_{t-3}$, which covers Cases~\hyperlink{caseI}{I} and~\hyperlink{caseIII}{III}.
\item[3.] $h_t = f_{t-4}$, which covers Cases~\hyperlink{caseI}{I} and~\hyperlink{caseII}{II}.
\end{itemize}

We initialize \cref{alg:IOMDA-DP,alg:OptIOMDA-DP} with random initial values since our analysis demonstrate that the upper bounds of the duality gap are independent of the initial values. 
The concern revolves around the growth pattern of the duality gap, particularly whether it exhibits linear or sublinear growth. 
To facilitate a clearer interpretation of this information from the image, we provide depictions of the time-averaged Dual-Gap. 
To ensure stability, we set $\epsilon=0.1$ to prevent an initial learning rate of infinity. 
Additionally, we apply the doubling trick to eliminate the dependence of our algorithms on $P$. 

\subsection{Results}

We run the repeated game for $10^5$ rounds and record time-averaged Dual-Gap curves. 
Results are shown in \cref{fig:results}, and all observed phenomena align with theoretical expectations.
For Case~\hyperlink{caseI}{I}, as shown in \cref{fig:case-i}, all average Dual-Gap curves converge, confirming that all algorithms asymptotically approximate saddle points. 
In the oscillating environment, specifically Case~\hyperlink{caseII}{II} (as shown in \cref{fig:case-ii}) and Case~\hyperlink{caseIII}{III} (as shown in \cref{fig:case-iii}), only OptIOMDA with $h_t=f_{t-4}$ and $h_t=f_{t-3}$ converge, respectively. 
OptIOMDA with $h_t=f_{t-4}$ asymptotically approaches saddle points in Case~\hyperlink{caseII}{II} because it covers scenarios where $\rho(f_t,f_{t-2})\rightarrow 0$. 
For adversarial OCCO of Case~\hyperlink{caseIV}{IV}, as illustrated in \cref{fig:case-iv}, all average Dual-Gap curves converge due to the sublinear growth of all $P_T$s. 

\begin{figure}[htbp]
\centering
\begin{subfigure}[H]{0.49\textwidth}
\caption{Average Dual-Gaps for Case \hyperlink{caseI}{I}}
\centering
\begin{tikzpicture}
\scriptsize
\begin{axis}[width=\textwidth,
             height=0.6\textwidth,
             xmode=log,xlabel={$T$},xlabel near ticks,
             ymode=log,ylabel={Average Dual-Gap},ylabel near ticks,
             mark=none,max space between ticks=10,
             axis x line*=none,
             axis y line*=none,
             xmin=1,
             xmax=100000,
             legend pos=north east,
             legend style={fill=none,
                           draw=none,
                           legend cell align=left},
            ]
\addplot[color=blue!70!cyan,smooth,very thick]
table[x=x,y=a,col sep=comma] {dualgap1.csv};
\addplot[color=green!60!black,smooth,very thick]
table[x=x,y=b,col sep=comma] {dualgap1.csv};
\addplot[color=red!65!orange,smooth,very thick]
table[x=x,y=c,col sep=comma] {dualgap1.csv};
\addplot[color=brown!60!olive,smooth,very thick]
table[x=x,y=d,col sep=comma] {dualgap1.csv};
\end{axis}
\end{tikzpicture}
\label{fig:case-i}
\end{subfigure}
\hfill
\begin{subfigure}[H]{0.49\textwidth}
\caption{Average Dual-Gaps for Case \hyperlink{caseII}{II}}
\centering
\begin{tikzpicture}
\scriptsize
\begin{axis}[width=\textwidth,
             height=0.6\textwidth,
             xmode=log,xlabel={$T$},xlabel near ticks,
             ymode=log,ylabel={Average Dual-Gap},ylabel near ticks,
             mark=none,max space between ticks=10,
             axis x line*=none,
             axis y line*=none,
             xmin=1,
             xmax=100000,
             legend pos=north east,
             legend style={fill=none,
                           draw=none,
                           legend cell align=left},
            ]
\addplot[color=blue!70!cyan,smooth,very thick]
table[x=x,y=a,col sep=comma] {dualgap2.csv};
\addplot[color=green!60!black,smooth,very thick]
table[x=x,y=b,col sep=comma] {dualgap2.csv};
\addplot[color=red!65!orange,smooth,very thick]
table[x=x,y=c,col sep=comma] {dualgap2.csv};
\addplot[color=brown!60!olive,smooth,very thick]
table[x=x,y=d,col sep=comma] {dualgap2.csv};
\end{axis}
\end{tikzpicture}
\label{fig:case-ii}
\end{subfigure}
\\[3pt]
\begin{subfigure}[H]{0.49\textwidth}
\caption{Average Dual-Gaps for Case \hyperlink{caseIII}{III}}
\centering
\begin{tikzpicture}
\scriptsize
\begin{axis}[width=\textwidth,
             height=0.6\textwidth,
             xmode=log,xlabel={$T$},xlabel near ticks,
             ymode=log,ylabel={Average Dual-Gap},ylabel near ticks,
             mark=none,max space between ticks=10,
             axis x line*=none,
             axis y line*=none,
             xmin=1,
             xmax=100000,
             legend pos=north east,
             legend style={fill=none,
                           draw=none,
                           legend cell align=left},
legend to name={legend:Dual-Gap},
legend style={at={(0.5,-0.2)},    
                    anchor=north,legend columns=4},]
\legend{IOMDA,OptIOMDA($h_t\!=\!f_{t\text{--}1}$),OptIOMDA($h_t\!=\!f_{t\text{--}3}$), OptIOMDA($h_t\!=\!f_{t\text{--}4}$)}
\addplot[color=blue!70!cyan,smooth,very thick]
table[x=x,y=a,col sep=comma] {dualgap3.csv};
\addplot[color=green!60!black,smooth,very thick]
table[x=x,y=b,col sep=comma] {dualgap3.csv};
\addplot[color=red!65!orange,smooth,very thick]
table[x=x,y=c,col sep=comma] {dualgap3.csv};
\addplot[color=brown!60!olive,smooth,very thick]
table[x=x,y=d,col sep=comma] {dualgap3.csv};
\end{axis}
\end{tikzpicture}
\label{fig:case-iii}
\end{subfigure}
\hfill
\begin{subfigure}[H]{0.49\textwidth}
\caption{Average Dual-Gaps for Case \hyperlink{caseIV}{IV}}
\centering
\begin{tikzpicture}
\scriptsize
\begin{axis}[width=\textwidth,
             height=0.6\textwidth,
             xmode=log,xlabel={$T$},xlabel near ticks,
             ymode=log,ylabel={Average Dual-Gap},ylabel near ticks,
             mark=none,max space between ticks=10,
             axis x line*=none,
             axis y line*=none,
             xmin=1,
             xmax=100000,
             legend pos=north east,
             legend style={fill=none,
                           draw=none,
                           legend cell align=left},
            ]
\addplot[color=blue!70!cyan,smooth,very thick]
table[x=x,y=a,col sep=comma] {dualgap4.csv};
\addplot[color=green!60!black,smooth,very thick]
table[x=x,y=b,col sep=comma] {dualgap4.csv};
\addplot[color=red!65!orange,smooth,very thick]
table[x=x,y=c,col sep=comma] {dualgap4.csv};
\addplot[color=brown!60!olive,smooth,very thick]
table[x=x,y=d,col sep=comma] {dualgap4.csv};
\end{axis}
\end{tikzpicture}
\label{fig:case-iv}
\end{subfigure}
\\[6pt]
{\scriptsize\hypersetup{linkcolor=black}\ref{legend:Dual-Gap}\hypersetup{linkcolor=darkred}}
\caption{Results of \cref{alg:IOMDA-DP,alg:OptIOMDA-DP}. }
\label{fig:results}
\end{figure}

\section{Conclusion and Discussion}

This work addresses the Online Saddle Point~(OSP) problem within the Online Convex-Concave Optimization~(OCCO) framework. 
We introduce the generalized duality gap~(Dual-Gap) as the performance metric and establish a parallel nature between OCCO with Dual-Gap and OCO with regret. 
This work deepens the understanding of OSP problems and their connections with OCCO and OCO.

In the OSP problem, the primary emphasis is on achieving convergence to Nash equilibria, where competitors correspond to saddle points. 
Consequently, this paper employs a dynamic observation of competitors. 
In fact, the OCCO framework allows arbitrary competitor sequences. 
Intuitively, OCCO should inherit the characteristics of OCO, but verifying this might not be straightforward. 
Below, we discuss the difficulties encountered when allowing for arbitrary competitors. 

Revisiting OCO, when the learning rate is determined by the preset upper limit $P$, two approaches emerge: 1) maintaining the assumption of arbitrariness of the competitor sequence and employing the meta-expert framework; 2) assuming the competitor sequence can be observed on the fly and applying the doubling trick for $P$. 
These two approaches share an intrinsic connection. 
Let's delve deeper into the details. 
In the meta-expert framework, a group of experts is utilized, where expert $i$ is assigned $P=2^i p$, and the meta-layer tracks the optimal expert. 
In the doubling trick, the gaming process is segmented into stages $1, 2, \cdots$, where the value of $P$ in stage $i$ is twice that of stage $i-1$. Transitioning to the next stage occurs when the path length exceeds the value of $P$ of the current stage. 
Consequently, the doubling trick switches stages as the path length increases, representing a dynamic pursuit of the accurately preset expert. 
Meanwhile, the essence of the meta-expert framework lies in combining experts' advice, which in the long term are as robust as any single expert's advice, to track the accurately preset expert. 

The theoretical foundation supporting the meta-expert framework lies in the decomposition of regret as follows: 
\begin{equation}
\label{eq:regret-decomp}
\textup{regret}\ u_{1:T}=\sum_{t=1}^T \Big(\ell_t\left(x_t\right)-\ell_t(x_t^i)\Big)+\sum_{t=1}^T \Big(\ell_t(x_t^i)-\ell_t\left(u_t\right)\Big).
\end{equation}
The second term in \cref{eq:regret-decomp} represents the regret specific to expert $i$, and the first term leads to the meta-regret: 
\begin{equation*}
\ell_t\left(x_t\right)-\ell_t(x_t^i)
\leqslant\left\langle L_t, w_t-1_i\right\rangle,
\end{equation*}
where $L_t=\big[\ell_t(x_t^j)\big]_j$ represents the loss function of the meta-layer. 
Unfortunately, it appears that OCCO does not support the transplantation of the meta-expert framework. 
Attempting to mimic the format of \cref{eq:regret-decomp} is not feasible. 
Although the Dual-Gap has the following decomposition form: 
\begin{equation*}
\begin{aligned}
\sum_{t=1}^T \Big(f_t\left(x_t, v_t\right)-f_t(x_t^i, v_t)\Big)+\sum_{t=1}^T\Big(f_t(x_t^i, v_t)-f_t(u_t, y_t^i)\Big)+\sum_{t=1}^T\Big(f_t(u_t, y_t^i)-f_t\left(u_t, y_t\right)\Big),
\end{aligned}
\end{equation*}
the meta-layer loss function cannot be determined because of the arbitrariness of competitors, that is, $f_t\left(x_t, v_t\right)-f_t(x_t^i, v_t)
\leqslant\langle L_t^x, w_t-1_i\rangle$, where $L_t^x=\big[f_t(x_t^j, v_t)\big]_j$ dependent on $v_t$. 
Indeed, we also tried the following decomposition form: 
\begin{subequations}
\begin{align}
f_t(x_t,v_t)-f_t(u_t, y_t)
=\ &f_t(x_t, v_t)-f_t(x_t,\widetilde{y}_{t+1}^i)+f_t(\widetilde{x}_{t+1}^i, y_t)-f_t(u_t, y_t) \label{eq:dg-term1}\\
&+f_t(x_t,\widetilde{y}_{t+1}^i)-f_t(x_t^i,\widetilde{y}_{t+1}^i)+f_t(\widetilde{x}_{t+1}^i,y_t^i)-f_t(\widetilde{x}_{t+1}^i, y_t) \label{eq:dg-term2}\\
&+f_t(x_t^i,\widetilde{y}_{t+1}^i)-h_t(x_t^i,\widetilde{y}_{t+1}^i)+h_t(\widetilde{x}_{t+1}^i,y_t^i)-f_t(\widetilde{x}_{t+1}^i,y_t^i) \label{eq:dg-term3}\\
&+h_t(x_t^i,\widetilde{y}_{t+1}^i)-h_t(x_t^i,y_t^i)+h_t(x_t^i,y_t^i)-h_t(\widetilde{x}_{t+1}^i,y_t^i). \label{eq:dg-term4}
\end{align}
\end{subequations}
Although the update rule can be derived from \cref{eq:dg-term1,eq:dg-term4}, and \cref{eq:dg-term3} effectively characterizes the prediction error, determining the loss function of the meta-layer from \cref{eq:dg-term2} remains challenging due to the term $(\widetilde{x}_{t+1}^i,\widetilde{y}_{t+1}^i)$ related to expert $i$.


\bibliography{reference}

\newpage
\appendix
\section{Discussion on NE-regret}
\label{app:NE-regret}

Why regret analysis is important for OCO: Sublinear regret implies approximate coarse correlated equilibria. 
Why duality gap is important for time-varying two-player zero-sum games: Sublinear duality gap implies that in most of the iterates the agents use strategies that are approximate NE for the particular time. 
Why dynamic NE-regret is important: It is not clear to me. 
Imagine that in half of the iterates, the value is much larger than the value at NE, while in the other half iterates the opposite. 
This result in a small dynamic NE-regret, but it wouldn't indicate how closely the value of each iterate approaches NE. 
Remark 2 of \citet{zhang2022noregret} does not hit the weakness of NE-regret. 
In fact, all the mentioned metrics relying on function values, while the shortcoming of NE-regret is exclusive to it. 

In this appendix, we continue to employ IOMDA and OptIOMDA as our base algorithms. 
We also configure exclusive adaptive learning rates for NE-regret and proceed to conduct experiments that emphasize the limitations of this metric. 
Detailed proofs for this section are provided in \cref{sec:proofs}. 

\subsection{NE-regret bound for IOMDA}

This part employs IOMDA as the basic algorithm and uses NE-regret as the performance metric to design adaptive learning rates. 
The following two propositions correspond to \cref{lem:IOMDA,thm:IOMDA}, respectively. 

\begin{proposition}
\label{lem:IOMDA-ne}
Let $B_\phi$ and $B_\psi$ be $L_\phi$-Lipschitz on $X$ and $L_\psi$-Lipschitz on $Y$ respectively w.r.t. their first variables, and let $D_\phi^2$ and $D_\psi^2$ be the supremum of $B_\phi$ and $B_\psi$, respectively. 
IOMDA with non-increasing learning rates guarantees that 
\begin{equation*}
\begin{aligned}
\textrm{NE-regret}_{T}
\leqslant \frac{D^2+L P_T^\infty}{\eta_T} + \max\left\{S_T^x, S_T^y\right\}. 
\end{aligned}
\end{equation*}
where $D=\max\{D_\phi, D_\psi\}$, $L=\max\{L_\phi, L_\psi\}$, $P_T^\infty= \sum_{t=1}^T\max\big\{\lVert x_t^*-x_{t-1}^*\rVert,\lVert y_t^*-y_{t-1}^*\rVert\big\}$, and 
\begin{equation}
\label{eq:S}
\begin{aligned}
S_T^x=&\textstyle \sum_{t=1}^{T}\big[f_t(x_t,y_t)-f_t(x_{t+1}, y_{t+1})-B_{\phi}\big(x_{t+1}, x_{t}^\phi\big)/\eta_t\big], \\
S_T^y=&\textstyle \sum_{t=1}^{T}\big[f_t(x_{t+1}, y_{t+1})-f_t(x_t,y_t)-B_{\psi}\big(y_{t+1}, y_t^\psi\big)/\eta_t\big].
\end{aligned}
\end{equation}
\end{proposition}

One can apply adaptive trick as employed in \citet[][Theorem~5.1]{Campolongo2021closer} to configure learning rates. 
However, in contrast to \cref{thm:IOMDA}, an additional assumption is required. 

\begin{assumption}
\label{ass:payoff-bounded}
All payoff functions are bounded by $M$, that is, 
$|f_t(x,y)|\leqslant M$, $\forall x\in X$, $\forall y\in Y$ and for all $t$. 
\end{assumption}

\begin{proposition}
\label{thm:IOMDA-ne}
Under the assumptions of \cref{lem:IOMDA-ne} and \cref{ass:gradient-bounded,ass:payoff-bounded}, let $P$ be a preset upper bound of $P_T^\infty$. 
IOMDA with $(D^2+L P)/\eta_t=\epsilon + \sum_{\tau=1}^{t-1}\delta_{\tau}$, $\delta_t=(\mathit{\Sigma}_t-\max_{\tau\in 1:t-1}\mathit{\Sigma}_{\tau})_+$ and $\mathit{\Sigma}_t=\max\{(S_t^x)_+, (S_t^y)_+\}$ guarantees that 
\begin{equation*}
\begin{aligned}
\textrm{NE-regret}_{T}\leqslant O\left(\min\bigg\{\sum_{t=1}^{T}\rho\left(f_{t}, f_{t-1}\right), \sqrt{(1+P)T}\bigg\}\right),
\end{aligned}
\end{equation*}
where $\rho\left(f_{t}, f_{t-1}\right)=\max_{x\in X,\,y\in Y}\left\lvert f_t(x,y)-f_{t-1}(x,y)\right\rvert$ measures the distance between $f_t$ and $f_{t-1}$, and the constant $\epsilon > 0$ prevents $\eta_1$ from being infinite. 
\end{proposition}

We compile the above method into \cref{alg:IOMDA-ne}. 

\begin{algorithm}[htb]
\caption{IOMDA with Adaptive Learning Rates Guaranteed by NE-regret}
\label{alg:IOMDA-ne}
\begin{algorithmic}[1]
\STATE {\bfseries Require:} $1$-strongly convex regularizers $\phi\colon\mathscr{X}\rightarrow\mathbb{R}$ and $\psi\colon\mathscr{Y}\rightarrow\mathbb{R}$ satisfying the assumptions of \cref{lem:IOMDA-ne}, and $\{f_t\}_{t\geqslant1}$ satisfying \cref{ass:gradient-bounded,ass:payoff-bounded}.
\STATE {\bfseries Initialize:} $P$, $t=0$, $\epsilon>0$, $D=\max\{D_\phi, D_\psi\}$, $L=\max\{L_\phi, L_\psi\}$.
\REPEAT
\STATE $t\gets t+1$.
\STATE Solve $\eta_t$ by $(D^2+L P)/\eta_t=\epsilon + \sum_{\tau=1}^{t-1}\delta_{\tau}$.
\STATE Output $\left(x_t,y_t\right)$, and then observe $f_t$.
\STATE $F_t(x,y)=f_t(x,y)+\frac{1}{\eta_t}B_{\phi}\big(x, x_t^\phi\big)-\frac{1}{\eta_t}B_{\psi}\big(y, y_t^\psi\big)$.
\STATE Update $x_{t+1}=\arg\min_{x\in X}\max_{y\in Y}F_t(x,y)$. 
\STATE Update $y_{t+1}=\arg\max_{y\in Y}\min_{x\in X}F_t(x,y)$.
\STATE Calculate $S_t^x$, $S_t^y$ according to \cref{eq:S}. 
\STATE Get $\delta_{t}$ following the setting of \cref{thm:IOMDA-ne}. 
\STATE Solve $x_{t}^*=\arg\min_{x\in X}\max_{y\in Y}f_t(x,y)$ and $y_{t}^*=\arg\max_{y\in Y}\min_{x\in X}f_t(x,y)$. 
\STATE $P_t^\infty\gets\sum_{\tau=1}^t\max\big\{\lVert x_\tau^*-x_{\tau-1}^*\rVert,\lVert y_\tau^*-y_{\tau-1}^*\rVert\big\}$.
\UNTIL{$P < P_t^\infty$}
\end{algorithmic}
\end{algorithm}

\subsection{NE-regret bound for OptIOMDA}

This part uses OptIOMDA as the foundational algorithm and utilizes NE-regret as the performance metric for designing adaptive learning rates. 
The following two propositions correspond to \cref{lem:OptIOMDA,thm:OptIOMDA}, respectively.

\begin{proposition}
\label{lem:OptIOMDA-ne}
Under the assumptions of \cref{lem:IOMDA-ne}, and let $h_{t}$ be an arbitrary predictor. 
OptIOMDA with non-increasing learning rates guarantees that 
\begin{equation*}
\begin{aligned}
\textrm{NE-regret}_{T}
\leqslant \frac{D^2+L P_T^\infty}{\eta_T} + \sum_{t=1}^{T} \delta_t. 
\end{aligned}
\end{equation*}
where $D=\max\{D_\phi, D_\psi\}$, $L=\max\{L_\phi, L_\psi\}$, $P_T^\infty= \sum_{t=1}^T\max\big\{\lVert x_t^*-x_{t-1}^*\rVert,\lVert y_t^*-y_{t-1}^*\rVert\big\}$, and $\delta_t=\max\{\delta_t^x, \delta_t^y\}\geqslant 0$, 
\begin{equation}
\label{eq:SS}
\begin{aligned}
\delta_t^x &=f_t(x_t,y_t)-h_t(x_t,y_t)+h_t(\widetilde{x}_{t+1},y_t)-f_t(\widetilde{x}_{t+1}, y_t)-B_{\phi}\big(\widetilde{x}_{t+1}, x_t^\phi\big)/\eta_t, \\
\delta_t^y &=h_t(x_t,y_t)-f_t(x_t,y_t)+f_t(x_t, \widetilde{y}_{t+1})-h_t(x_t,\widetilde{y}_{t+1})-B_{\psi}\big(\widetilde{y}_{t+1}, y_t^\psi\big)\eta_t. \\
\end{aligned}
\end{equation}
\end{proposition}

Note that $\delta_t\geqslant 0$, so one can use adaptive trick adopted in \citet[][Theorem~5.1]{Campolongo2021closer} to set learning rates. 

\begin{proposition}
\label{thm:OptIOMDA-ne}
Under the assumptions of \cref{lem:OptIOMDA-ne} and let $f_t$ and $h_t$ satisfy \cref{ass:gradient-bounded}, let $P$ be a preset upper bound of $P_T^\infty$. 
OptIOMDA with $(D^2+L P)/\eta_t=\epsilon + \sum_{\tau=1}^{t-1}\delta_{\tau}$, $\delta_t=\max\{\delta_t^x, \delta_t^y\}$ incurs 
\begin{equation*}
\begin{aligned}
\textrm{NE-regret}_{T}\leqslant O\left(\min\bigg\{\sum_{t=1}^{T}\rho\left(f_{t}, h_{t}\right), \sqrt{(1+P)T}\bigg\}\right),
\end{aligned}
\end{equation*}
where the constant $\epsilon > 0$ prevents $\eta_1$ from being infinite. 
\end{proposition}

We compile the above method into \cref{alg:OptIOMDA-ne}. 

\begin{algorithm}[htb]
\caption{OptIOMDA with Adaptive Learning Rates Guaranteed by NE-regret}
\label{alg:OptIOMDA-ne}
\begin{algorithmic}[1]
\STATE {\bfseries Require:} $1$-strongly convex regularizers $\phi\colon\mathscr{X}\rightarrow\mathbb{R}$ and $\psi\colon\mathscr{Y}\rightarrow\mathbb{R}$ satisfying the assumptions of \cref{lem:OptIOMDA-ne}, $\{f_t\}_{t\geqslant1}$ and predictor sequence $\{h_t\}_{t\geqslant1}$ satisfying \cref{ass:gradient-bounded}. 
\STATE {\bfseries Initialize:} $P$, $t=0$, $\epsilon>0$, $D=\max\{D_\phi, D_\psi\}$, $L=\max\{L_\phi, L_\psi\}$.
\REPEAT
\STATE $t\gets t+1$.
\STATE Solve $\eta_t$ by $(D^2+L P)/\eta_t=\epsilon + \sum_{\tau=1}^{t-1}\delta_{\tau}$.
\STATE Receive $h_t$, and let $H_t(x,y)=h_t(x,y)+\frac{1}{\eta_t}B_{\phi}\big(x, \widetilde{x}_{t}^\phi\big)-\frac{1}{\eta_t}B_{\psi}\big(y, \widetilde{y}_{t}^\psi\big)$.
\STATE Update $x_{t}=\arg\min_{x\in X}\max_{y\in Y}H_t(x,y)$. 
\STATE Update $y_{t}=\arg\max_{y\in Y}\min_{x\in X}H_t(x,y)$.
\STATE Output $\left(x_t,y_t\right)$, and then observe $f_t$.
\STATE Update $\widetilde{x}_{t+1}=\arg\min_{x\in X}f_t(x,y_t)+\frac{1}{\eta_t}B_{\phi}\big(x, \widetilde{x}_{t}^\phi\big)$.
\STATE Update $\widetilde{y}_{t+1}=\arg\max_{y\in Y}f_t(x_t,y)-\frac{1}{\eta_t}B_{\psi}\big(y, \widetilde{y}_{t}^\psi\big)$.
\STATE Calculate $\delta_t^x$ and $\delta_t^y$ according to \cref{eq:SS}.
\STATE $\delta_t\gets\max\{\delta_t^x, \delta_t^y\}$.
\STATE Solve $x_{t}^*=\arg\min_{x\in X}\max_{y\in Y}f_t(x,y)$ and $y_{t}^*=\arg\max_{y\in Y}\min_{x\in X}f_t(x,y)$. 
\STATE $P_t^\infty\gets\sum_{\tau=1}^t\max\big\{\lVert x_\tau^*-x_{\tau-1}^*\rVert,\lVert y_\tau^*-y_{\tau-1}^*\rVert\big\}$.
\UNTIL{$P < P_t^\infty$}
\end{algorithmic}
\end{algorithm}

\subsection{Experiments}
\label{app:sec:Experiments}

This part provides experimental evidence highlighting the inherent limitations of NE-regret as a performance metric. 
We closely adhere to the experimental settings outlined in \cref{sec:setup}. 
It is noteworthy that NE-regret restricts competitors to the saddle-point sequence. 
Hence, in this part, we exclude Case~\hyperlink{caseIV}{IV}. 
%
To emphasize the point that the convergence of average NE-regret doesn't necessarily ensure the algorithm's ability to truly approach saddle points, we introduce the concept of time-averaged tracking error, which is given by 
\begin{equation*}
\begin{aligned}
\frac{1}{T}\sum_{t=1}^T\max\{\lVert x_t-x_t^*\rVert, \lVert y_t-y_t^*\rVert\}. 
\end{aligned}
\end{equation*}

\begin{figure}[tb]
\centering
\begin{subfigure}[H]{0.49\textwidth}
\caption{Average NE-regrets for Case \hyperlink{caseI}{I}}
\centering
\begin{tikzpicture}
\scriptsize
\begin{axis}[width=\textwidth,
             height=0.6\textwidth,
             xmode=log,xlabel={$T$},xlabel near ticks,
             ymode=log,ylabel={Average NE-regret},ylabel near ticks,
             mark=none,max space between ticks=10,
             axis x line*=none,
             axis y line*=none,
             xmin=1,
             xmax=100000,
             legend pos=north east,
             legend style={fill=none,
                           draw=none,
                           legend cell align=left},
            ]
\addplot[color=blue!70!cyan,smooth,very thick]
table[x=x,y=a,col sep=comma] {NE1.csv};
\addplot[color=green!60!black,smooth,very thick]
table[x=x,y=b,col sep=comma] {NE1.csv};
\addplot[color=red!65!orange,smooth,very thick]
table[x=x,y=c,col sep=comma] {NE1.csv};
\addplot[color=brown!60!olive,smooth,very thick]
table[x=x,y=d,col sep=comma] {NE1.csv};
\end{axis}
\end{tikzpicture}
\label{fig:case11}
\end{subfigure}
\hfill
\begin{subfigure}[H]{0.49\textwidth}
\caption{Average Tracking Errors for Case \hyperlink{caseI}{I}}
\centering
\begin{tikzpicture}
\scriptsize
\begin{axis}[width=\textwidth,
             height=0.6\textwidth,
             xmode=log,xlabel={$T$},xlabel near ticks,
             ymode=log,ylabel={Average Error},ylabel near ticks,
             mark=none,max space between ticks=8,
             axis x line*=none,
             axis y line*=none,
             xmin=1,
             xmax=100000,
             legend pos=north east,
             legend style={fill=none,
                           draw=none,
                           legend cell align=left},
            ]
\addplot[color=blue!70!cyan,smooth,very thick]
table[x=x,y=a,col sep=comma] {ERR1.csv};
\addplot[color=green!60!black,smooth,very thick]
table[x=x,y=b,col sep=comma] {ERR1.csv};
\addplot[color=red!65!orange,smooth,very thick]
table[x=x,y=c,col sep=comma] {ERR1.csv};
\addplot[color=brown!60!olive,smooth,very thick]
table[x=x,y=d,col sep=comma] {ERR1.csv};
\end{axis}
\end{tikzpicture}
\label{fig:case12}
\end{subfigure}
\\[3pt]
\begin{subfigure}[H]{0.49\textwidth}
\caption{Average NE-regrets for Case \hyperlink{caseII}{II}}
\centering
\begin{tikzpicture}
\scriptsize
\begin{axis}[width=\textwidth,
             height=0.6\textwidth,
             xmode=log,xlabel={$T$},xlabel near ticks,
             ymode=log,ylabel={Average NE-regret},ylabel near ticks,
             mark=none,max space between ticks=10,
             axis x line*=none,
             axis y line*=none,
             xmin=1,
             xmax=100000,
             legend pos=north east,
             legend style={fill=none,
                           draw=none,
                           legend cell align=left},
            ]
\addplot[color=blue!70!cyan,smooth,very thick]
table[x=x,y=a,col sep=comma] {NE2.csv};
\addplot[color=green!60!black,smooth,very thick]
table[x=x,y=b,col sep=comma] {NE2.csv};
\addplot[color=red!65!orange,smooth,very thick]
table[x=x,y=c,col sep=comma] {NE2.csv};
\addplot[color=brown!60!olive,smooth,very thick]
table[x=x,y=d,col sep=comma] {NE2.csv};
\end{axis}
\end{tikzpicture}
\label{fig:case21}
\end{subfigure}
\hfill
\begin{subfigure}[H]{0.49\textwidth}
\caption{Average Tracking Errors for Case \hyperlink{caseII}{II}}
\centering
\begin{tikzpicture}
\scriptsize
\begin{axis}[width=\textwidth,
             height=0.6\textwidth,
             xmode=log,xlabel={$T$},xlabel near ticks,
             ymode=log,ylabel={Average Error},ylabel near ticks,
             mark=none,max space between ticks=10,
             axis x line*=none,
             axis y line*=none,
             xmin=1,
             xmax=100000,
             legend pos=north east,
             legend style={fill=none,
                           draw=none,
                           legend cell align=left},
            ]
\addplot[color=blue!70!cyan,smooth,very thick]
table[x=x,y=a,col sep=comma] {ERR2.csv};
\addplot[color=green!60!black,smooth,very thick]
table[x=x,y=b,col sep=comma] {ERR2.csv};
\addplot[color=red!65!orange,smooth,very thick]
table[x=x,y=c,col sep=comma] {ERR2.csv};
\addplot[color=brown!60!olive,smooth,very thick]
table[x=x,y=d,col sep=comma] {ERR2.csv};
\end{axis}
\end{tikzpicture}
\label{fig:case22}
\end{subfigure}
\\[3pt]
\begin{subfigure}[H]{0.49\textwidth}
\caption{Average NE-regrets for Case \hyperlink{caseIII}{III}}
\centering
\begin{tikzpicture}
\scriptsize
\begin{axis}[width=\textwidth,
             height=0.6\textwidth,
             xmode=log,xlabel={$T$},xlabel near ticks,
             ymode=log,ylabel={Average NE-regret},ylabel near ticks,
             mark=none,max space between ticks=10,
             axis x line*=none,
             axis y line*=none,
             xmin=1,
             xmax=100000,
             legend pos=north east,
             legend style={fill=none,
                           draw=none,
                           legend cell align=left},
legend to name={legend:NE-regret},
legend style={at={(0.5,-0.2)},    
                    anchor=north,legend columns=4},]
\legend{IOMDA,OptIOMDA($h_t\!=\!f_{t\text{--}1}$),OptIOMDA($h_t\!=\!f_{t\text{--}3}$), OptIOMDA($h_t\!=\!f_{t\text{--}4}$)}
\addplot[color=blue!70!cyan,smooth,very thick]
table[x=x,y=a,col sep=comma] {NE3.csv};
\addplot[color=green!60!black,smooth,very thick]
table[x=x,y=b,col sep=comma] {NE3.csv};
\addplot[color=red!65!orange,smooth,very thick]
table[x=x,y=c,col sep=comma] {NE3.csv};
\addplot[color=brown!60!olive,smooth,very thick]
table[x=x,y=d,col sep=comma] {NE3.csv};
\end{axis}
\end{tikzpicture}
\label{fig:case31}
\end{subfigure}
\hfill
\begin{subfigure}[H]{0.49\textwidth}
\caption{Average Tracking Errors for Case \hyperlink{caseIII}{III}}
\centering
\begin{tikzpicture}
\scriptsize
\begin{axis}[width=\textwidth,
             height=0.6\textwidth,
             xmode=log,xlabel={$T$},xlabel near ticks,
             ymode=log,ylabel={Average Error},ylabel near ticks,
             mark=none,max space between ticks=7,
             axis x line*=none,
             axis y line*=none,
             xmin=1,
             xmax=100000,
             legend pos=north east,
             legend style={fill=none,
                           draw=none,
                           legend cell align=left},
            ]
\addplot[color=blue!70!cyan,smooth,very thick]
table[x=x,y=a,col sep=comma] {ERR3.csv};
\addplot[color=green!60!black,smooth,very thick]
table[x=x,y=b,col sep=comma] {ERR3.csv};
\addplot[color=red!65!orange,smooth,very thick]
table[x=x,y=c,col sep=comma] {ERR3.csv};
\addplot[color=brown!60!olive,smooth,very thick]
table[x=x,y=d,col sep=comma] {ERR3.csv};
\end{axis}
\end{tikzpicture}
\label{fig:case32}
\end{subfigure}
\\[6pt]
{\scriptsize\hypersetup{linkcolor=black}\ref{legend:NE-regret}\hypersetup{linkcolor=darkred}}
\caption{Average NE-regrets and Average Tracking Errors of \cref{alg:IOMDA-ne,alg:OptIOMDA-ne}. }
\label{fig:errors}
\end{figure}
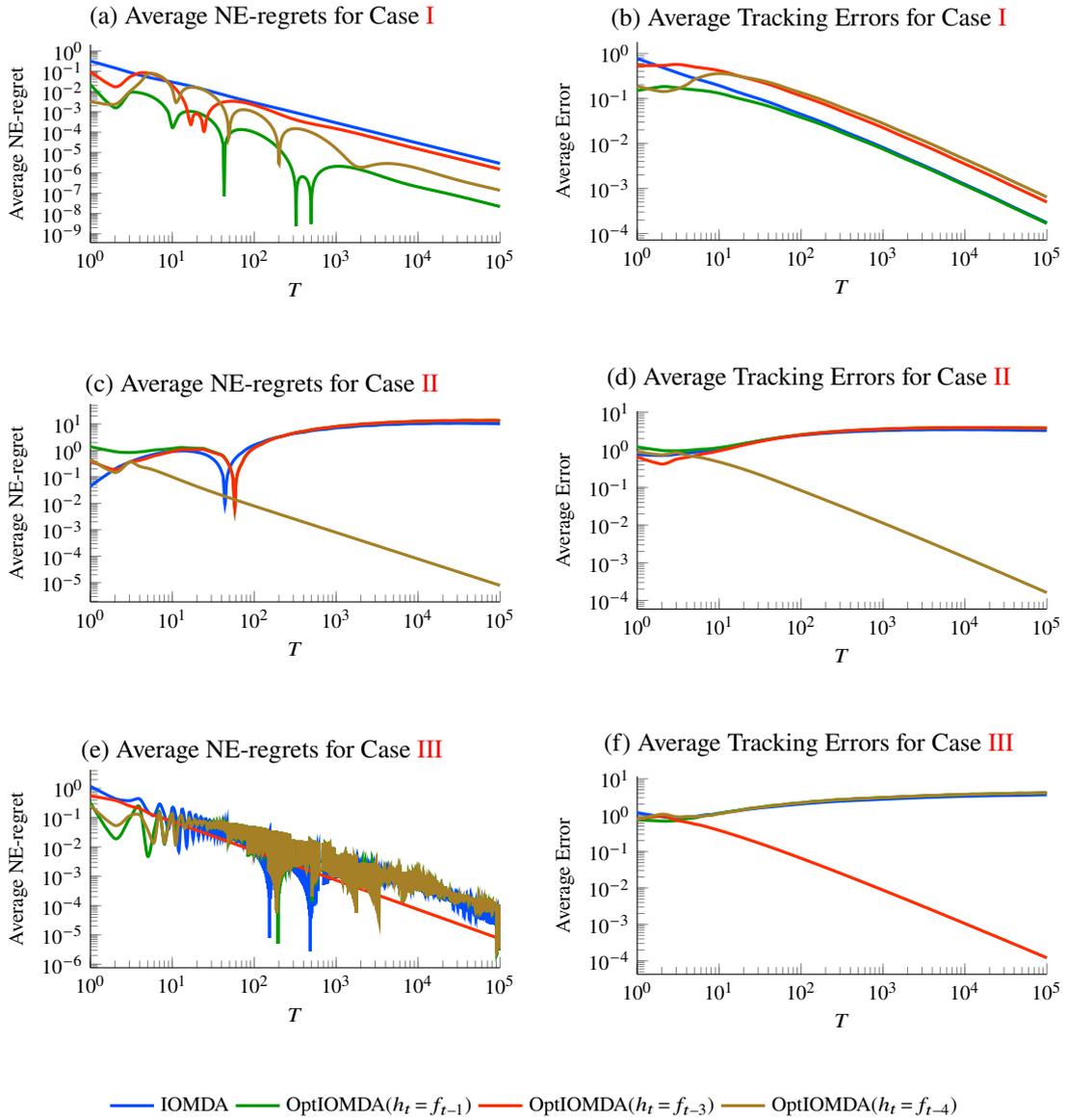

Experimental results are summarized in \cref{fig:errors}.
Let's examine the results in the oscillatory environment of Case~\hyperlink{caseIII}{III}. Although all average NE-regrets converge, the average tracking errors reveal that there are still instances where saddle points cannot be effectively tracked.
Indeed, only the outputs of OptIOMDA with $h_t=f_{t-3}$ asymptotically approach the saddle points. 
The violent oscillations in the other average NE-regret curves suggest cancellation occurring within the modulus operation, highlighting an inherent flaw of NE-regret. 

\subsection{Proofs}
\label{sec:proofs}

This part lists the proofs of \cref{lem:IOMDA-ne,thm:IOMDA-ne,lem:OptIOMDA-ne,thm:OptIOMDA-ne}. 
The analysis framework follows \citet[][Appendix~A]{Campolongo2021closer}. 

\begin{proof}[Proof of \cref{lem:IOMDA-ne}]
Applying convexity and first-order optimality condition yields the existence of $x_{t+1}^{f_t(\,\cdot\,, y_{t+1})}$, such that 
\begin{equation*}
\begin{aligned}
f_t(x_{t+1}, y_{t+1})-f_t(x_t^*, y_{t+1})
&\leqslant\big\langle x_{t+1}^{f_t(\,\cdot\,, y_{t+1})}, x_{t+1}-x_t^*\big\rangle 
\leqslant\big\langle x_t^\phi-x_{t+1}^\phi, x_{t+1}-x_t^*\big\rangle/\eta_t \\
&=\big[B_{\phi}\big(x_t^*, x_{t}^\phi\big)-B_{\phi}\big(x_t^*, x_{t+1}^\phi\big)\big]/\eta_t 
-B_{\phi}\big(x_{t+1}, x_{t}^\phi\big)/\eta_t.
\end{aligned}
\end{equation*}
Note that $f_t(x_t^*, y_{t+1})-f_t(x_t^*, y_t^*)\leqslant 0$ since $(x_t^*, y_t^*)$ is the saddle point of $f_t$. 
Combining all above inequalities yields 
\begin{equation*}
\begin{aligned}
f_t(x_t,y_t)-f_t(x_t^*, y_t^*) 
\leqslant\ &\big[B_{\phi}\big(x_t^*, x_{t}^\phi\big)-B_{\phi}\big(x_t^*, x_{t+1}^\phi\big)\big]/\eta_t \\
&+ f_t(x_t,y_t)-f_t(x_{t+1}, y_{t+1})-B_{\phi}\big(x_{t+1}, x_{t}^\phi\big)/\eta_t. 
\end{aligned}
\end{equation*}
Sum over time, we get 
\begin{equation*}
\begin{aligned}
\sum_{t=1}^{T}f_t(x_t,y_t) -\sum_{t=1}^{T} f_t(x_t^*, y_t^*)
\leqslant\underbrace{\sum_{t=1}^{T}\frac{1}{\eta_t}\big[B_{\phi}\big(x_t^*, x_{t}^\phi\big)-B_{\phi}\big(x_t^*, x_{t+1}^\phi\big)\big]}_{\hypertarget{terma}{\text{term a}}} 
+S_T^x,
\end{aligned}
\end{equation*}
where 
\begin{equation*}
\begin{aligned}
\textrm{term~\hyperlink{terma}{a}}
&\leqslant\sum_{t=1}^{T}\frac{1}{\eta_t}\left[B_{\phi}\big(x_t^*, x_{t}^\phi\big)-B_{\phi}\big(x_{t-1}^*, x_{t}^\phi\big)\right] 
+\frac{B_{\phi}\big(x_{0}^*, x_{1}^\phi\big)}{\eta_0} \\
&\hspace*{1.2em}+\sum_{t=1}^{T}\left(\frac{1}{\eta_{t}}-\frac{1}{\eta_{t-1}}\right)B_{\phi}\big(x_{t-1}^*, x_{t}^\phi\big) 
\leqslant\frac{D_\phi^2}{\eta_T}+\sum_{t=1}^{T}\frac{L_\phi}{\eta_t}\left\lVert x_t^*-x_{t-1}^*\right\rVert
\end{aligned}
\end{equation*}
since $\eta_t$ is non-increasing over time, $B_\phi$ is $L_\phi$-Lipschitz w.r.t. the first variable, and $D_\phi^2$ is the supremum of $B_\phi$. 
Similarly, applying concavity yields 
\begin{equation*}
\begin{aligned}
\sum_{t=1}^{T}f_t(x_t^*, y_t^*)-\sum_{t=1}^{T} f_t(x_t,y_t) \leqslant\frac{D_\psi^2}{\eta_T}+\sum_{t=1}^{T}\frac{L_\psi}{\eta_t}\left\lVert y_t^*-y_{t-1}^*\right\rVert
+S_T^y.
\end{aligned}
\end{equation*}
Let $D=\max\{D_\phi, D_\psi\}$, $L=\max\{L_\phi, L_\psi\}$, and let $P_T^\infty= \sum_{t=1}^T\max\big\{\lVert x_t^*-x_{t-1}^*\rVert,\lVert y_t^*-y_{t-1}^*\rVert\big\}$. 
Therefore, we get 
\begin{equation*}
\textrm{NE-regret}_{T}
=\left\lvert \sum_{t=1}^T f_t\left(x_t, y_t\right)-\sum_{t=1}^T f_t\left(x_t^*, y_t^*\right)\right\rvert 
\leqslant \frac{D^2+L P_T}{\eta_T} + \max\left\{S_T^x, S_T^y\right\}. \qedhere
\end{equation*}
\end{proof}

\begin{proof}[Proof of \cref{thm:IOMDA-ne}]
First, we claim that $\sum_{\tau=1}^t \delta_\tau = \max_{\tau\in 1:t}\mathit{\Sigma}_{\tau}$. 
This claim can be proved by induction. 
It is obvious that $\delta_1=\mathit{\Sigma}_1$. 
Now we assume the claim holds for $t-1$ and prove it for $t$: 
\begin{equation*}
\begin{aligned}
\sum_{\tau=1}^t \delta_\tau
=\delta_t + \sum_{\tau=1}^{t-1} \delta_\tau 
=\Big(\mathit{\Sigma}_t-\max_{\tau\in 1:t-1}\mathit{\Sigma}_{\tau}\Big)_+ + \max_{\tau\in 1:t-1}\mathit{\Sigma}_{\tau} 
= \max_{\tau\in 1:t}\mathit{\Sigma}_{\tau}.
\end{aligned}
\end{equation*}
Applying the prescribed learning rate $\eta_t$ , we rewrite the bound in \cref{lem:IOMDA-ne} as follows: 
\begin{equation*}
\begin{aligned}
\textrm{NE-regret}_{T}
\leqslant \epsilon+\sum_{t=1}^{T-1} \delta_t + \max_{t\in 1:T}\mathit{\Sigma}_{t} 
\leqslant \epsilon+2\sum_{t=1}^{T} \delta_t 
= \epsilon+2\max_{t\in 1:T}\mathit{\Sigma}_{t}. 
\end{aligned}
\end{equation*}
Next, we estimate the upper bound of the r.h.s. of the above inequality in two ways. 
The firse one: 
\begin{equation}
\label{eq:the-first-bound}
\begin{aligned}
\max_{t\in 1:T}\mathit{\Sigma}_{t}\leqslant 2M+\sum_{t=1}^{T}\rho\left( f_{t},f_{t-1}\right). 
\end{aligned}
\end{equation}
Indeed, the monotonicity of $(\,\cdot\,)_+$ implies that 
\begin{equation*}
\begin{aligned}
\mathit{\Sigma}_{t}^x 
&\leqslant\bigg(\sum_{\tau=1}^{t}\big[f_\tau(x_\tau,y_\tau)-f_\tau(x_{\tau+1}, y_{\tau+1})\big]\bigg)_+ \\
&\leqslant\big|f_{0}(x_1,y_1)-f_{t}(x_{t+1}, y_{t+1})\big| + \sum_{\tau=1}^{t}\big|f_{\tau}(x_\tau,y_\tau)-f_{\tau-1}(x_{\tau}, y_{\tau})\big| \\
&\leqslant 2M +\sum_{\tau=1}^{t}\rho\left( f_{\tau},f_{\tau-1}\right),
\end{aligned}
\end{equation*}
and similarly, $\mathit{\Sigma}_{t}^y\leqslant 2M +\sum_{\tau=1}^{t}\rho\left( f_{\tau},f_{\tau-1}\right)$. 
So \cref{eq:the-first-bound} holds. 
The other one: 
\begin{equation}
\label{eq:the-second-bound}
\begin{aligned}
\sum_{t=1}^{T}\delta_t\leqslant G\sqrt{\left(11 D^2+3 L P\right) T}. 
\end{aligned}
\end{equation}
The derivation is as follows. 
\begin{equation*}
\begin{aligned}
\delta_t
=\Big(\mathit{\Sigma}_t-\max_{\tau\in 1:t-1}\mathit{\Sigma}_{\tau}\Big)_+
\leqslant(\mathit{\Sigma}_t-\mathit{\Sigma}_{t-1})_+ 
\leqslant\begin{cases}
(\mathit{\Sigma}_t^x-\mathit{\Sigma}_{t-1}^x)_+, & \mathit{\Sigma}_t^x \geqslant \mathit{\Sigma}_t^y, \\
(\mathit{\Sigma}_t^y-\mathit{\Sigma}_{t-1}^y)_+, & \mathit{\Sigma}_t^x < \mathit{\Sigma}_t^y.
\end{cases}
\end{aligned}
\end{equation*}
For the first piecewise, 
\begin{equation*}
\begin{aligned}
(\mathit{\Sigma}_t^x-\mathit{\Sigma}_{t-1}^x)_+ 
&\leqslant\big(f_t(x_t,y_t)-f_t(x_{t+1}, y_{t+1})-B_{\phi}\big(x_{t+1}, x_{t}^\phi\big)/\eta_t\big)_+ \\
&\leqslant\underbrace{f_t(x_{t}, y_{t+1})-f_t(x_{t+1}, y_{t+1})-B_{\phi}\big(x_{t+1}, x_{t}^\phi\big)/\eta_t}_{\hypertarget{termb}{\text{term b}}} \\
&\hspace*{1.2em}+ \underbrace{\big| f_t(x_t,y_t)-f_t(x_{t}, y_{t+1})\big|}_{\hypertarget{termc}{\text{term c}}},
\end{aligned}
\end{equation*}
where the first ``$\leqslant$'' follows from a result which says that given $a,b\in\mathbb{R}$, $((a+b)_+-(a)_+)_+\leqslant(b)_+$ holds since $(\,\cdot\,)\leqslant(\,\cdot\,)_+$, and the last ``$\leqslant$'' holds since $\text{term \hyperlink{termb}{b}}\geqslant 0$, which can be obtained by applying $F_t(x_{t+1}, y_{t+1})\leqslant F_t(x_{t}, y_{t+1})$. 
Note that 
\begin{equation*}
\begin{aligned}
\text{term \hyperlink{termb}{b}}
&\leqslant G_X \left\lVert x_t-x_{t+1}\right\rVert-B_{\phi}\big(x_{t+1}, x_{t}^\phi\big)/\eta_t 
\leqslant G_X\sqrt{2B_{\phi}\big(x_{t+1}, x_{t}^\phi\big)}-B_{\phi}\big(x_{t+1}, x_{t}^\phi\big)/\eta_t \\
&\leqslant\min\big\{\sqrt{2}D_\phi G_X,\,\eta_t G_X^2/2\big\},
\end{aligned}
\end{equation*}
and term~\hyperlink{termc}{c} $\leqslant G_Y \left\lVert y_t-y_{t+1}\right\rVert\leqslant\min\{\sqrt{2}D_\psi G_Y, \eta_t G_Y^2\}$, where the first term in the minimum follows from the $1$-strong-convexity of $\psi$ and the boundedness of $B_\psi$, i.e., $D_\psi^2\geqslant B_{\psi}\big(y_{t+1}, y_{t}^\psi\big)\geqslant\frac{1}{2}\left\lVert y_t-y_{t+1}\right\rVert^2$, 
and the second term in the minimum follows from the $\eta_t^{-1}$-strong-convexity of $\varphi=-F_t(x_{t+1},\,\cdot\,)$. 
Indeed, $y_{t+1}^\varphi=0$, 
\begin{equation*}
\begin{aligned}
B_{\varphi}\big(y_{t+1}, y_{t}^\varphi\big)\geqslant\textstyle \frac{1}{2\eta_t}\left\lVert y_t-y_{t+1}\right\rVert^2, \qquad
B_{\varphi}\big(y_{t}, y_{t+1}^\varphi\big)\geqslant\textstyle \frac{1}{2\eta_t}\left\lVert y_t-y_{t+1}\right\rVert^2.
\end{aligned}
\end{equation*}
Adding the above two inequalities and applying the equality condition of Fenchel-Young inequality yields $\left\lVert y_t-y_{t+1}\right\rVert\leqslant \eta_t \left\lVert y_{t}^\varphi\right\rVert=\eta_t \left\lVert \partial_y(-f_t)(x_{t+1},y_t) \right\rVert\leqslant\eta_t G_Y$. 
Now we get 
\begin{equation*}
\begin{aligned}
(\mathit{\Sigma}_t^x-\mathit{\Sigma}_{t-1}^x)_+ 
\leqslant\min\big\{\sqrt{2}\big(D_\phi G_X+D_\psi G_Y\big),\,\eta_t \big(G_X^2/2+G_Y^2\big)\big\}.
\end{aligned}
\end{equation*}
In a similar way, for the second piecewise, 
\begin{equation*}
\begin{aligned}
(\mathit{\Sigma}_t^y-\mathit{\Sigma}_{t-1}^y)_+ 
\leqslant\min\big\{\sqrt{2}\big(D_\phi G_X+D_\psi G_Y\big),\,\eta_t \big(G_Y^2/2+G_X^2\big)\big\}.
\end{aligned}
\end{equation*}
Thus, $\delta_t\leqslant\min\big\{2\sqrt{2}DG,\,3\eta_t G^2/2\big\}$. 
Now we have that 
\begin{equation*}
\begin{aligned}
\left(\sum_{t=1}^{T}\delta_t\right)^2
&=\sum_{t=1}^{T}\delta_t^2+2\sum_{t=1}^{T}\delta_t\sum_{\tau=1}^{t-1}\delta_\tau 
=\sum_{t=1}^{T}\delta_t^2+2\sum_{t=1}^{T}\delta_t \left(\frac{D^2+L P}{\eta_t}-\epsilon\right) \\
&\leqslant \sum_{t=1}^{T}8 D^2 G^2+\sum_{t=1}^{T} 3G^2(D^2+L P),
\end{aligned}
\end{equation*}
where the inequality uses the first and second terms in the minimum of the bound for $\delta_t$ in turn. 
So \cref{eq:the-second-bound} holds. 
In conclusion, 
\begin{equation*}
\textrm{NE-regret}_{T}\leqslant \epsilon + 
2\min\left\{2M+\sum_{t=1}^{T}\rho\left( f_{t},f_{t-1}\right),\ G\sqrt{\left(11 D^2+3 L P\right) T}\right\}. \qedhere
\end{equation*}
\end{proof}

\begin{proof}[Proof of \cref{lem:OptIOMDA-ne}]
On the one hand, 
\begin{equation*}
\begin{aligned}
f_t(x_t,y_t)&-f_t(x_t^*, y_t^*) 
=f_t(x_t,y_t)-h_t(x_t,y_t)+h_t(\widetilde{x}_{t+1},y_t)-f_t(\widetilde{x}_{t+1}, y_t) \\
&+\underbrace{f_t(x_t^*, y_t)-f_t(x_t^*, y_t^*)}_{\hypertarget{termA}{\text{term A}}} 
+\underbrace{f_t(\widetilde{x}_{t+1}, y_t)-f_t(x_t^*, y_t)}_{\hypertarget{termB}{\text{term B}}}+\underbrace{h_t(x_t,y_t)-h_t(\widetilde{x}_{t+1},y_t)}_{\hypertarget{termC}{\text{term C}}},
\end{aligned}
\end{equation*}
where $\text{term \hyperlink{termA}{A}}\leqslant 0$, and according to first-order optimality condition, $\exists \widetilde{x}_{t+1}^{f_t(\,\cdot\,, y_t)}$ and $x_t^{h_t(\,\cdot\,,y_t)}$, such that 
\begin{equation*}
\begin{aligned}
\text{term \hyperlink{termB}{B}}&\leqslant\big\langle\widetilde{x}_{t+1}^{f_t(\,\cdot\,, y_t)}, \widetilde{x}_{t+1}-x_t^*\big\rangle 
\leqslant\big\langle \widetilde{x}_{t}^\phi-\widetilde{x}_{t+1}^\phi,\widetilde{x}_{t+1}-x_t^*\big\rangle/\eta_t \\
&=\big[B_{\phi}\big(x_t^*, \widetilde{x}_{t}^\phi\big)-B_{\phi}\big(x_t^*, \widetilde{x}_{t+1}^\phi\big)-B_{\phi}\big(\widetilde{x}_{t+1}, \widetilde{x}_{t}^\phi\big)\big]/\eta_t, \\
\text{term \hyperlink{termC}{C}}&\leqslant\big\langle x_t^{h_t(\,\cdot\,,y_t)}, x_t-\widetilde{x}_{t+1}\big\rangle 
\leqslant\big\langle \widetilde{x}_{t}^\phi-x_{t}^\phi,x_t-\widetilde{x}_{t+1}\big\rangle /\eta_t \\
&=\big[B_{\phi}\big(\widetilde{x}_{t+1}, \widetilde{x}_t^\phi\big)-B_{\phi}\big(\widetilde{x}_{t+1}, x_t^\phi\big)-B_{\phi}\big(x_t, \widetilde{x}_t^\phi\big)\big]/\eta_t.
\end{aligned}
\end{equation*}
After rearranging, we get
\begin{equation*}
\begin{aligned}
f_t(x_t,y_t)-f_t(x_t^*, y_t^*) 
\leqslant\big[B_{\phi}\big(x_t^*, \widetilde{x}_{t}^\phi\big)-B_{\phi}\big(x_t^*, \widetilde{x}_{t+1}^\phi\big)\big]/\eta_t+\delta^x_t.
\end{aligned}
\end{equation*}
On the other hand, in a similar way, we get 
\begin{equation*}
\begin{aligned}
f_t(x_t,y_t)-f_t(x_t^*, y_t^*) 
\geqslant-\big[B_{\psi}\big(y_t^*, \widetilde{y}_{t}^\psi\big)-B_{\psi}\big(y_t^*, \widetilde{y}_{t+1}^\psi\big)\big]/\eta_t-\delta^y_t.
\end{aligned}
\end{equation*}
Using the same trick as term~\hyperlink{terma}{a} yields 
\begin{equation*}
\begin{aligned}
\textrm{NE-regret}_{T} 
=\left\lvert \sum_{t=1}^T f_t\left(x_t, y_t\right)-\sum_{t=1}^T f_t\left(x_t^*, y_t^*\right)\right\rvert 
\leqslant \frac{D^2+L P_T^\infty}{\eta_T} + \sum_{t=1}^{T} \delta_t,
\end{aligned}
\end{equation*}
where $D=\max\{D_\phi, D_\psi\}$, $L=\max\{L_\phi, L_\psi\}$, $P_T^\infty= \sum_{t=1}^T\max\big\{\lVert x_t^*-x_{t-1}^*\rVert,\lVert y_t^*-y_{t-1}^*\rVert\big\}$, and $\delta_t=\max\{\delta_t^x, \delta_t^y\}$. 
Now let's verify $\delta_t\geqslant 0$. 
Indeed, it suffices to verify $\delta_t^x\geqslant 0$ and $\delta_t^y\geqslant 0$. 
According to the optimality of $\widetilde{x}_{t+1}$, we get 
\begin{equation*}
\begin{aligned}
f_t(x_t,y_t)+\frac{1}{\eta_t}B_{\phi}\big(x_t, \widetilde{x}_{t}^\phi\big)\geqslant f_t(\widetilde{x}_{t+1},y_t)+\frac{1}{\eta_t}B_{\phi}\big(\widetilde{x}_{t+1}, \widetilde{x}_{t}^\phi\big).
\end{aligned}
\end{equation*}
The inequality related to term~\hyperlink{termC}{C} says 
\begin{equation*}
\begin{aligned}
-h_t(x_t,y_t)+h_t(\widetilde{x}_{t+1},y_t)
\geqslant -\frac{1}{\eta_t}\left[B_{\phi}\big(\widetilde{x}_{t+1}, \widetilde{x}_t^\phi\big)-B_{\phi}\big(\widetilde{x}_{t+1}, x_t^\phi\big)-B_{\phi}\big(x_t, \widetilde{x}_t^\phi\big)\right].
\end{aligned}
\end{equation*}
Combining the above two inequalities immediately yields $\delta_t^x\geqslant 0$. 
$\delta_t^y\geqslant 0$ can be obtained in a similar way. 
\end{proof}

\begin{proof}[Proof of \cref{thm:OptIOMDA-ne}]
Note that $(D^2+L P)/\eta_t=\epsilon+\sum_{\tau=1}^{t-1}\delta_{\tau}$, where $P$ is an upper bound of the path length $P_T^\infty$, and $\epsilon>0$ is a constant to prevent $\eta_1$ from being infinite. 
Such a learning rate setting guarantees that 
\begin{equation*}
\begin{aligned}
\textrm{NE-regret}_{T}
\leqslant \epsilon+2\sum_{t=1}^{T} \delta_t. 
\end{aligned}
\end{equation*}
We provide two upper bounds for $\delta_t$. 
The first one: 
\begin{equation*}
\begin{aligned}
\delta_t=\max\{\delta_t^x, \delta_t^y\}\leqslant 2\max_{x\in X,\,y\in Y}\lvert f_t(x,y)-h_t(x,y)\rvert=2\rho\left(f_t, h_t\right).
\end{aligned}
\end{equation*}
The second one: 
\begin{equation*}
\begin{aligned}
\delta_t\leqslant 2\min\left\{\sqrt{2}D G,\,\eta_t G^2\right\}.
\end{aligned}
\end{equation*}
Indeed, 
\begin{equation*}
\begin{aligned}
\delta_t^x
&\leqslant\big\langle \partial_x f_t(x_t, y_t)-\partial_x h_t(\widetilde{x}_{t+1}, y_t), x_t-\widetilde{x}_{t+1}\big\rangle-B_{\phi}\big(\widetilde{x}_{t+1}, x_t^\phi\big)/\eta_t \\
&\leqslant\big\lVert \partial_x f_t(x_t, y_t)-\partial_x h_t(\widetilde{x}_{t+1}, y_t)\big\rVert\sqrt{2 B_{\phi}\big(\widetilde{x}_{t+1}, x_t^\phi\big)}-B_{\phi}\big(\widetilde{x}_{t+1}, x_t^\phi\big)/\eta_t \\
&\leqslant 2\min\left\{\sqrt{2}D_\phi G_X,\,\eta_t G_X^2\right\},
\end{aligned}
\end{equation*}
and similarly, $\delta_t^y\leqslant 2\min\left\{\sqrt{2}D_\psi G_Y,\,\eta_t G_Y^2\right\}$. 
Sum over time for the two cases separately, we get 
\begin{equation*}
\begin{aligned}
\sum_{t=1}^{T}\delta_t\leqslant 2\sum_{t=1}^{T}\rho\left(\ell_t, h_t\right),
\end{aligned}
\end{equation*}
and 
\begin{equation*}
\begin{aligned}
\left(\sum_{t=1}^{T}\delta_t\right)^2
&=\sum_{t=1}^{T}\delta_t^2+2\sum_{t=1}^{T}\delta_t\sum_{\tau=1}^{t-1}\delta_\tau=\sum_{t=1}^{T}\delta_t^2+2\sum_{t=1}^{T}\delta_t\left(\frac{D^2+L P}{\eta_t}-\epsilon\right) \\
&\leqslant \sum_{t=1}^{T}8 D^2 G^2+\sum_{t=1}^{T}4G^2(D^2+L P)
= 4G^2(3D^2+L P)T. 
\end{aligned}
\end{equation*}
In conclusion, 
\begin{equation*}
\textrm{NE-regret}_{T}\leqslant \epsilon+4\min\left\{\sum_{t=1}^{T}\rho\left(f_t, h_t\right),\ G\sqrt{(3D^2+L P)T}\right\}. 
\qedhere
\end{equation*}
\end{proof}

\end{document}